%% file: neurips_2026.tex
\documentclass{article}

 \usepackage[main, final]{neurips_2026}

\usepackage[utf8]{inputenc} %
\usepackage[T1]{fontenc}    %
\usepackage[hypertexnames=false]{hyperref}       %
\usepackage{url}            %
\usepackage{booktabs}       %
\usepackage{amsfonts}       %
\usepackage{nicefrac}       %
\usepackage{microtype}      %

\usepackage{amsmath}
\usepackage[capitalize,noabbrev]{cleveref}
\usepackage{multirow}

\usepackage{bbding}
\usepackage{tabularx}
\usepackage{graphicx}
\usepackage[table,xcdraw]{xcolor}
\usepackage{color}
\definecolor{minetable1colorx}{rgb}{0.75, 0.75, 0.75}

\definecolor{Yellow}{RGB}{176, 224, 230}

\usepackage{enumitem}

\newtheorem{theorem}{Theorem}[section]
\AddToHook{env/theorem/begin}{\crefalias{section}{theorem}}
\usepackage{algorithm}		%
\usepackage{algorithmic}	%
\usepackage{caption}

\title{Ultra-DPO: Reimagining LLM Alignment\\as a Semi-Supervised Task}

\author{Li He\textsuperscript{1,2} \quad Jiaxian Guo\textsuperscript{3} \quad Yaqing Wang\textsuperscript{4}
\quad Yue Yang\textsuperscript{5} \quad He Zhao\textsuperscript{2} \\ \textbf{Stephen Wan\textsuperscript{2}   \quad Dadong Wang\textsuperscript{2} \quad Lina Yao\textsuperscript{6} \quad Tongliang Liu\textsuperscript{1}}\thanks{Corresponding authors.} \\
\textsuperscript{1} The University of Sydney \quad \textsuperscript{2} CSIRO, Data61 \quad \textsuperscript{3} Google Research \\
\textsuperscript{4} Beijing Institute of Mathematical Sciences and Applications \\
\textsuperscript{5} Maincode \quad \textsuperscript{6} University of New South Wales\\
}

\begin{document}

\maketitle
\input{0_Abstract}
\input{1_Introduction}
\input{2_Related_Work}
\input{3_Preliminary}
\input{4_Theoretical}
\input{5_Method}
\input{6_Experiment}

\input{7_Conclusion}

\bibliography{references}
\bibliographystyle{plainnat}

\input{8_Appendix}

\newpage
\input{checklist.tex}

\end{document}

%% file: 0_Abstract.tex
\begin{abstract}

    Direct Preference Optimization (DPO) offers a streamlined one-stage alternative to Reinforcement Learning from Human Feedback (RLHF). However, our theoretical analysis reveals a structural distinction between the two: DPO and RLHF optimize alignment on different data distributions with different supervision sources. Specifically, DPO learns from ground-truth human preferences but within the off-policy support of an offline dataset, while RLHF optimizes on the on-policy rollouts but learns from pseudo-labels produced by a reward proxy. This distinction explains how each method may fail: DPO alignment collapses when the policy drifts away from the offline dataset support, and RLHF alignment is biased due to the extrapolation error of the reward proxy on out-of-distribution rollouts. To mitigate both failure modes, we propose a semi-supervised alignment objective: ground-truth preferences anchor alignment within the dataset support, and pseudo-labels on online rollouts extend it beyond. Applying DPO's reward reparameterization to this objective yields Ultra-DPO, a one-stage algorithm that augments direct human alignment with an online self-alignment mechanism. Across multiple benchmarks, Ultra-DPO consistently outperforms DPO, RLHF, and online preference learning methods. Our implementation is available at \href{https://github.com/heli-stand/Ultra-DPO}{\texttt{github.com/heli-stand/Ultra-DPO}}.

\end{abstract}

%% file: 1_Introduction.tex
\section{Introduction}
Aligning large language models (LLMs) \citep{abdin2025phi, olmo2025olmo, adcock2026llama, liu2026ministral} with human preferences has emerged as a central problem in modern post-training, with two dominant paradigms: Reinforcement Learning from Human Feedback (RLHF) \citep{casper2023open, kaufmann2024survey} and Direct Preference Optimization (DPO) \citep{rafailov2023direct}. RLHF takes a two-stage approach: a reward model is first trained on offline human preferences, and the LLM is then optimized against this learned reward proxy via online RL. DPO bypasses the reward modeling stage entirely, allowing LLMs to be optimized directly on the offline preference dataset. DPO is widely regarded as a streamlined one-stage alternative to RLHF, eliminating the complexity of reward model training and online RL while reportedly achieving comparable alignment quality.

\begin{figure}[t]
    \centering
    \includegraphics[width=1\linewidth, clip, trim = 9mm 39mm 25mm 30mm]{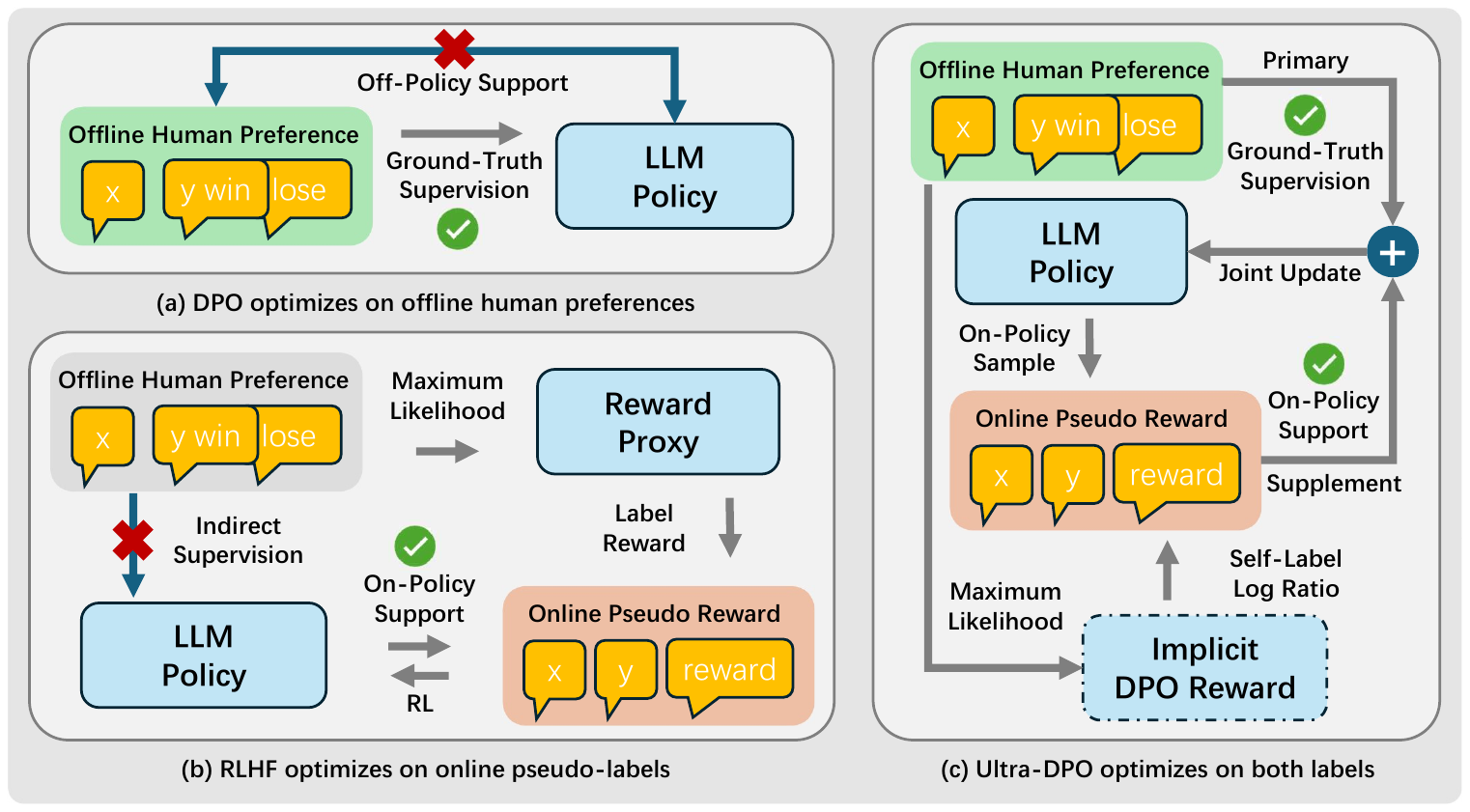}
    \vspace{1mm}
    \caption{\textbf{Alignment paradigms compared.} \textbf{(a) DPO} directly aligns from ground-truth human preferences but is restricted to the off-policy support of a fixed offline dataset; alignment collapses when the policy drifts away from this support. \textbf{(b) RLHF} optimizes on on-policy rollouts but against pseudo-labels from a learned reward proxy; alignment is biased by the proxy's extrapolation error on out-of-distribution rollouts. \textbf{(c) Ultra-DPO (ours)} jointly optimizes on offline human preferences and online pseudo-labels in a single stage to mitigate both failure modes. Ground-truth supervision anchors alignment to limit the effects of reward-proxy bias, while on-policy pseudo-labels extend alignment beyond the offline support to mitigate distribution mismatch.}
    \label{fig:ultra_dpo_main}
\end{figure}

However, this conventional framing on the procedural differences obscures a structural distinction between the two methods. Specifically, our analysis reveals that DPO and RLHF optimize alignment on different data distributions with different supervision sources. As shown in \cref{fig:ultra_dpo_main}(a), DPO learns from ground-truth human preferences \citep{stiennon2020learning, wang2024helpsteer, ji2025pku}, but with the off-policy support of a fixed offline dataset. By contrast, RLHF optimizes on the on-policy rollouts from the current policy \citep{song2024importance, tajwar2024preference}, but learns from pseudo-labels produced by the learned reward proxy, as illustrated in \cref{fig:ultra_dpo_main}(b). Such a structural distinction implies that both methods may fail during the alignment process. On one hand, DPO alignment collapses as the policy drifts away from the offline dataset support, because the increasing divergence between data and policy compromises the monotonic improvement guarantee \citep{schulman2017trustregionpolicyoptimization}. On the other hand, RLHF alignment is biased by the extrapolation error of the reward proxy \citep{laidlaw2023preventing, dai2025mitigating}, because the proxy is queried on online rollouts that lie increasingly further from its training distribution. Effective alignment therefore requires addressing both limitations simultaneously, which motivates a new alignment paradigm.

To this end, we propose a semi-supervised approach to LLM alignment as shown in \cref{fig:ultra_dpo_main}(c). The insight is to treat ground-truth human preferences as the primary signal and on-policy reward labels as a supplement. The former anchors alignment within the human preference support, helping limit the effects of reward misestimation on out-of-distribution rollouts; the latter extends alignment to on-policy rollouts, helping mitigate the distribution mismatch associated with off-policy collapse. Applying DPO's reward reparameterization to this semi-supervised objective yields Ultra-DPO, a one-stage algorithm that augments offline preference alignment with an online self-alignment mechanism. Ultra-DPO addresses both structural failure modes identified above while retaining DPO's practical advantage of avoiding explicit reward model training. Without additional computational overhead compared to a standard RLHF pipeline, our approach outperforms a comprehensive set of preference learning and RLHF baselines.
We summarize our contributions as follows:
\begin{itemize}[leftmargin=0.4cm,topsep=-2pt]
\item We identify a structural distinction between DPO and RLHF along optimization distribution and supervision source. This predicts a failure mode for each method, and we verify both empirically.
\item We propose a semi-supervised alignment objective that addresses both failure modes. From this objective, we derive Ultra-DPO, a one-stage alignment algorithm that augments DPO with an online self-alignment mechanism.
\item We conduct comprehensive experiments showing that Ultra-DPO consistently outperforms preference learning methods (DPO, IPO, SimPO) and RLHF methods (PPO, RLOO, GRPO, DAPO).
\end{itemize}

%% file: 2_Related_Work.tex
\section{Related Work}

\textbf{Direct Alignment Algorithms.} Several offline direct alignment algorithms (DAAs) have been proposed to address specific limitations of DPO. IPO \citep{azar2024general} reframes the objective to prevent overfitting to the implicit reward model through a regularized loss formulation, while KTO~\citep{ethayarajh2024kto} relaxes the pairwise preference requirement by learning from individual binary feedback signals. CPO~\citep{xu2024contrastive} and SimPO \citep{meng2024simpo} eliminate the reference model dependency by assuming a uniform reference policy, and ORPO \citep{hong2024orpo} further simplifies alignment by removing the separate SFT stage. To address limitations of offline alignment, online (or iterative) DAAs \citep{guo2024direct} leverage on-policy preference labels to guide policy updates beyond static offline datasets. These methods typically implement preference judges using pre-trained reward models \citep{chen2024optune,xu2024dpo} or verifiable reward functions~\citep{pang2024iterative, tu2025enhancing}. While our method targets the same limitation as online DAAs, we achieve this without external supervision from reward models or generative LLM feedback.

\textbf{Self-Aligning LLMs.} The reliance on extensive human annotations for alignment presents significant scalability challenges, motivating research into self-alignment methods where models autonomously improve with minimal human supervision. One prominent approach uses synthetic data generation, where models create their own SFT datasets through self-critique and revision based on predefined principles \citep{bai2022constitutional} or self-generated instruction-response pairs \citep{wang2023self}. Beyond SFT, another line of work focuses on creating preference datasets using self-scoring prompts \citep{yuan2024self, wu2025meta} or self-generated rankings \citep{wang2024cream,lu2025urpo}. Additionally, research on weak-to-strong generalization \citep{burns2023weak} explores settings where weaker models supervise stronger ones, offering insights into scalable oversight. The most closely related work is by \citet{chen2024bootstrapping}, which proposes a self-alignment stage following standard post-training that uses self-derived implicit DPO rewards. This work validates that implicit rewards learned during DPO contain sufficient signal to guide further self-alignment, demonstrating that preference learning itself generates reusable supervision. Building on this insight, our method differs by integrating self-alignment directly into the preference learning stage, enabling LLMs to learn from self-generated DPO rewards alongside human preferences. Consequently, our approach achieves one-stage RLHF without requiring additional post-training phases.

%% file: 3_Preliminary.tex
\section{Preliminary}

This section establishes the mathematical foundation for our theoretical analysis and proposed method. We first review the two-stage RLHF pipeline: reward modeling and preference alignment. We then present DPO and its derivation, as well as online DPO, where the expectation is computed over preferences from iteratively collected on-policy samples.

\textbf{Reward Modeling.} In the task of LLM alignment, we first try to capture human preferences by learning a reward model $r_\phi$ out from a binary preference dataset $\mathcal{D}$, where a response $y_w$ is more preferable over another response $y_l$ given the same prompt $x$. By framing it as a logistic regression problem using the Bradley-Terry model \citep{bradley1952rank}, we derive the optimal reward model by optimizing the log-likelihood objective:
\begin{align}\label{eq:reward_model}
    \mathcal{J}_\text{RM}(r_\phi)=\mathbb{E}_{(x, y_w, y_l)\sim \mathcal{D}(x,y_w, y_l)} \bigl[\log \sigma(r_{\phi}(x, y_w)- r_{\phi}(x, y_l))\bigr],
\end{align}

\textbf{Preference Alignment.} In the second stage, we train a language model $\pi_\theta$ to output reward-maximizing responses based on the trained reward model $r_\phi$. This is achieved via maximizing the reward expectation with a reference regularization \citep{ziegler2019fine, stiennon2020learning}:
\begin{align}
\label{eq:RL}
\mathcal{J}_\text{RLHF-Align}(\pi_\theta)=\mathbb{E}_{x\sim \mathcal{D}(x), y\sim \pi_\theta(\cdot|x)} \Bigl[r_{\phi}(x, y) - \beta\mathbb{D}_{\textrm{KL}}\bigl[\pi_{\theta}(\cdot|x) || \pi_\text{ref}(\cdot|x)\bigr]\Bigr],
\end{align}
where $\mathcal{D}(x)$ is the prompt dataset, $\pi_\text{ref}$ is the reference policy, $\beta$ is the regularization coefficient. The KL divergence term prevents the policy from moving far from the reference distribution, mitigating the reward hacking issue \citep{eisenstein2023helping,gao2023scaling, rafailov2024scaling}.

\textbf{DPO.} \citet{rafailov2023direct} propose a two-in-one preference alignment algorithm directly based on offline human preferences. The derivation starts from the observation that KL-constrained RL objectives like \cref{eq:RL} have closed-form solutions, following previous works \citep{peters2007reinforcement,peng2019advantage, korbak2022reinforcement, go2023aligning}. This gives us a way to reparameterize the reward model using the learning policy $\pi_\theta$ and the reference policy $\pi_\text{ref}$:
\begin{equation}\label{eq:reparameterization}
    r_\phi(x,y)=\beta \log{\frac{\pi_{\theta}(y|x)}{\pi_\text{ref}(y|x)}}+\beta \log Z(x), %
\end{equation}
where $Z(x)=\sum_y \pi_\text{ref}(y|x)\exp\Big(\frac{1}{\beta} r_\phi(x,y)\Big)$ is the partition function ensuring the solution policy is a valid probability distribution. By substituting \cref{eq:reparameterization} in \cref{eq:reward_model}, we derive the DPO objective in terms of only the learning policy for direct human preference alignment:
\begin{align}\label{eq:dpo}
    &\mathcal{J}_\text{DPO}(\pi_{\theta}) = \mathbb{E}_{(x, y_w, y_l)\sim\mathcal{D}(x,y_w, y_l)}  \left[\log \sigma \left(\beta \log \frac{\pi_{\theta}(y_w|x)}{\pi_\text{ref}(y_w|x)} - \beta \log \frac{\pi_{\theta}(y_l|x)}{\pi_\text{ref}(y_l|x)}\right)\right].
\end{align}

\textbf{Online DPO.} To address the off-policy issue of standard DPO, online DPO was proposed \citep{xiong2023iterative, guo2024direct}, which replaces the offline dataset with a dynamically constructed dataset $\mathcal{D}_\text{Online}(x,y_w, y_l)$. Such a dataset is constructed via a combined procedure of (1) two-shot generation of $(y_1, y_2)\sim \pi_\theta(\cdot|x)$, and (2) online preference labeling $(y_w, y_l) \sim p_{\text{judge}}(y_1, y_2|x)$ using a preference judge $p_\text{judge}$.  A pre-trained reward model via \cref{eq:reward_model} often serves as the judge, the chosen response $y_w$ is determined as the one with the higher reward score.

%% file: 4_Theoretical.tex
\section{What DPO and RLHF Actually Optimize}\label{sec:theoretical}
DPO and RLHF are conventionally distinguished by whether reward modeling is explicit. We argue this framing overlooks a deeper structural distinction. The two methods differ along two axes: 1) supervision source, which determines whether feedback comes directly from human annotations or from a learned reward model; 2) optimization distribution, which determines whether feedback is applied to fixed offline comparisons or current-policy generations. This distinction highlights two limitations: improvements on offline comparisons may not transfer to generated responses, while feedback on current generations depends on reward-model generalization. We develop this perspective in \cref{sec:dpo_offline,sec:rlhf_online} and examine its empirical implications in \cref{sec:validation}. Appendix~\ref{app:extended_theoreticals} further discusses the relationship between online DPO and RLHF within this framework.

\subsection{DPO: Ground Truth, but Off-Policy}\label{sec:dpo_offline}
DPO optimizes \cref{eq:dpo} as a supervised loss over offline preference pairs. While prior work has analyzed properties of DPO's optimization landscape \citep{xu2024dpo}, its implicit alignment objective has not been characterized. We characterize it as below with detailed proof in Appendix~\ref{proof:dpo_implicit}:
\begin{theorem}
\label{thm:dpo_implicit}
Let $\mathcal{D}$ be a preference dataset over tuples $(x, y_w, y_l)$, and assume $\beta>0$, $\pi_\theta(y|x) > 0$ and $\pi_\textnormal{ref}(y|x) > 0$ for all $(x,y)$ in $\mathcal{D}$.
Then DPO can be interpreted as optimizing the Bradley-Terry objective whose sample-wise reward margin is induced by the within-pair optimality condition of the following alignment objective:
\begin{align}\label{eq:implicit_align}
  \mathcal{J}_\textnormal{DPO-Align}(\pi_\theta)
    = \mathbb{E}_{(x,y_w,y_l)\sim\mathcal{D}(x, y_w, y_l)} \left[\sum_{y\in\{y_w,y_l\}}
    \pi_\theta(y|x)\left(
      r_\phi(x,y)
      - \beta\log\frac{\pi_\theta(y|x)}{\pi_\textnormal{ref}(y|x)}
    \right)\right].
\end{align}
\end{theorem}

The DPO's implicit alignment objective sees ground-truth labels, but only on a support that is fixed offline and disjoint from where $\pi_\theta$ actually places its mass. The inner sum ranges only over the two completions labeled by humans, because the reward reparameterization is partial: $r_\phi(x,y)$ is identified only on response pairs with preference labels. Within the labeled pair, however, DPO's supervision is reliable: $r_\phi(x,y)$ is fit on exactly these in-distribution responses.

The structural failure mode follows directly. Taking improvement steps on the offline distribution violates the policy improvement guarantee \citep{kakade2002approximately, schulman2017trustregionpolicyoptimization}, which require the data distribution to remain close to the policy being improved. As the estimated improvement direction diverges from the true one, alignment collapses on the policy's actual rollouts.

\subsection{RLHF: On-Policy, but Pseudo-Labeled}\label{sec:rlhf_online}
To make the structural comparison with DPO, we rewrite the RLHF alignment objective in \cref{eq:RL} as a sum over the on-policy support, with derivation in Appendix~\ref{proof:rlhf_sum}:
\begin{align} \label{eq:RL_expectation_sum}
    \mathcal{J}_\textnormal{RLHF-Align}(\pi_\theta) = \mathbb{E}_{x\sim \mathcal{D}(x)}  \left[\sum_{y\in \textnormal{supp}(\pi_\theta(\cdot| x))} \pi_\theta(y| x)\left(r_\phi(x,y) - \beta\log \frac{\pi_\theta(y|x)}{\pi_\textnormal{ref}(y|x)} \right)\right].
\end{align}

\cref{eq:implicit_align} and \ref{eq:RL_expectation_sum} share the same KL-regularized objective and differ only along the two axes of our framework. On the optimization axis, the inner sum ranges over the on-policy support rather than the labeled pair $\{y_w, y_l\}$, so RLHF optimizes alignment on the policy's actual behaviors and thus avoids DPO's off-policy collapse. On the supervision axis, however, the reward model is queried on online rollouts that increasingly lie outside its training support, as $\pi_\theta$ drifts from $\pi_\text{ref}$.

This mismatch makes RLHF alignment biased by the reward-proxy extrapolation error. The reliability of $r_\phi$ degrades with distance from its training distribution \citep{gao2023scaling, dai2025mitigating}, so on-policy rollouts that drift from this support receive biased reward estimates. Optimizing against this biased proxy causes $\pi_\theta$ to improve under $r_\phi$ while diverging from ground-truth alignment.

\subsection{Empirical Validation}\label{sec:validation}
To examine the predicted failure modes, we run controlled experiments on HH-RLHF, using GRPO as the RLHF implementation. Full experimental details are in Appendix~\ref{app:validation}.

\begin{figure*}[t]
    \centering
    {\includegraphics[scale=0.83]{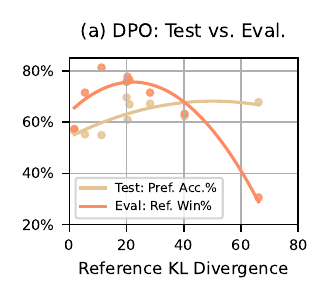} \hspace{-2mm}\
        \includegraphics[scale=0.83]{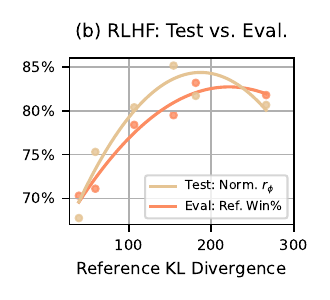} \hspace{-2mm}\
        \includegraphics[scale=0.83]{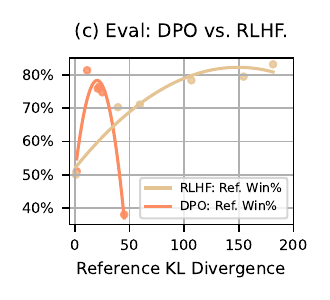}
        }
    \vspace{-3mm}
    \caption{Empirical comparison of offline DPO and GRPO-based RLHF on HH-RLHF.
    (a) DPO's held-out preference accuracy improves while its win rate against the reference policy peaks and then declines.
    (b) For RLHF, the learned reward score and win rate improve together over the evaluated training range.
    (c) In the low-KL regime, DPO achieves higher observed win rates at comparable KL.
    KL is measured relative to $\pi_{\textnormal{ref}}$.}
    \label{fig:empirical}
\end{figure*}

\textbf{DPO: preference fitting and generation quality diverge.}
\Cref{fig:empirical}(a) shows that DPO's preference accuracy on held-out pairs continues to improve as KL increases, while its win rate against $\pi_{\textnormal{ref}}$ peaks and then declines. This pattern demonstrates that better generalization to offline comparisons may not translate into better generations. It is consistent with the off-policy generalization gap identified.

\textbf{RLHF: reward and win rate improve together.}
In \cref{fig:empirical}(b), the learned reward score and win rate improve together over the evaluated training range. Thus, the decline observed for DPO does not occur in this RLHF run. Joint improvement remains compatible with imperfect reward feedback: reward-model error can limit the gains achieved without producing an absolute decline in win rate.

\textbf{Performance at comparable KL.}
We further compare the methods in the low-KL regime, before the decline observed for DPO. \Cref{fig:empirical}(c) shows that DPO achieves higher observed win rates at comparable KL, giving a better KL--win-rate trade-off in this regime. This result is consistent with a benefit from direct preference supervision.

These observations motivate retaining direct supervision from human preference pairs while incorporating reward feedback on current-policy generations. The resulting combination can connect observed preferences to the behaviors that emerge as the policy evolves.

%% file: 5_Method.tex
\section{Ultra-DPO}\label{sec:method}
After identifying the limitations of existing alignment approaches, our goal is to propose a new alignment objective that addresses both failure modes for more effective LLM alignment. To this end, we propose a semi-supervised objective that optimizes the policy on both offline human preferences and on-policy rewards. Our key insight is to use ground-truth preferences as the primary signal to limit the effects of reward-proxy bias, with on-policy reward labels providing a supplementary signal to mitigate the distribution mismatch associated with off-policy collapse. From this objective, we derive Ultra-DPO, a policy-only one-stage algorithm that augments direct preference alignment with an online self-alignment mechanism. With appropriate weighting and regularization, the resulting update can mitigate both failure modes without explicit reward model training.

\textbf{Derivation.} We formulate the Ultra-DPO objective by jointly optimizing reward modeling and semi-supervised alignment. Combining the reward modeling objective in \cref{eq:reward_model} with the DPO and RLHF alignment objectives in \cref{eq:implicit_align,eq:RL}, we define:
\begin{align}
    \mathcal{J}_\text{Ultra-DPO}(r_\phi, \pi_\theta)=\mathcal{J}_\text{RM}(r_\phi)+ \underbrace{\mathcal{J}_\text{DPO-Align}(\pi_\theta)+\lambda\cdot\mathcal{J}_\text{RLHF-Align}(\pi_\theta)}_\text{$\mathcal{J}_\text{Semi-Align}$: Semi-Supervised Alignment}, \nonumber
\end{align}
where $\lambda>0$ controls the weight of the supplementary on-policy signal. Let $\beta_1$ and $\beta_2$ be
the KL coefficients of the DPO and RLHF alignment terms respectively, we expand $\mathcal{J}_\text{Semi-Align}$ and obtain:
\begin{align} \label{eq:udpo_expand}
& \mathcal{J}_\text{Ultra-DPO}(r_\phi, \pi_\theta)=\mathcal{J}_\text{RM}(r_\phi) +\lambda \cdot\mathbb{E}_{x\sim \mathcal{D}(x), y\sim \pi_\theta(\cdot|x)} \Bigl[\text{sg}(r_{\phi}(x, y)) - \beta_2\mathbb{D}_{\textrm{KL}}\bigl[\pi_{\theta} \| \pi_\text{ref}\bigr]\Bigr]+\nonumber
\\& \qquad \mathbb{E}_{(x, y_w, y_l)\sim \mathcal{D}(x, y_w, y_l)}\left[ \sum_{y \in \{y_w, y_l\}} \pi_\theta(y|x)\left(\text{sg}(r_{\phi}(x, y)) - \beta_1 \log \frac{\pi_{\theta}(y|x)}{\pi_\textnormal{ref}(y|x)}  \right)\right]
\end{align}
where $\text{sg}(\cdot)$ denotes the stop-gradient operator, which prevents gradients from backpropagating through reward labeling and ensures the semi-supervised objective updates only the policy model.

For sufficiently small $\lambda$, we approximate the solution to
$\mathcal{J}_{\text{Semi-Align}}$ by that of
$\mathcal{J}_{\text{DPO-Align}}$. Using the offline alignment term's
within-pair optimality condition, we apply DPO's reward
reparameterization in \cref{eq:reparameterization} to eliminate
the offline alignment subproblem and express the reward-modeling
likelihood $\mathcal{J}_{\text{RM}}$ as the standard DPO objective
$\mathcal{J}_{\text{DPO}}$. This gives the reduced formulation
\begin{align}
    \mathcal{J}_{\text{Ultra-DPO}}(r_\phi,\pi_\theta)
    :={}& \mathcal{J}_{\text{DPO}}(\pi_\theta)
    + \lambda \cdot
    \mathbb{E}_{x\sim\mathcal{D}(x),\,
    y\sim\pi_\theta(\cdot|x)}
    \Biggl[
        \text{sg}\bigl(r_\phi(x,y)\bigr)
        - \beta_2
        \log\frac{\pi_\theta(y| x)}
        {\pi_{\text{ref}}(y|x)}
    \Biggr].
    \nonumber
\end{align}
Optimizing $\mathcal{J}_{\text{DPO}}$ fits the pairwise margins of
the implicit reward
$\beta_1\log\bigl(\pi_\theta(y|x)/
\pi_{\text{ref}}(y|x)\bigr)$
to the observed preference labels. We reuse this learned score
as a detached pseudo-reward on current-policy generations,
replacing $r_\phi$ in the on-policy term. The resulting Ultra-DPO
objective depends only on the policy and requires no explicit
reward model:
\begin{align}
    \mathcal{J}_{\text{Ultra-DPO}}(\pi_\theta)
    :={}& \mathcal{J}_{\text{DPO}}(\pi_\theta)
    + \lambda \cdot
    \mathbb{E}_{x\sim\mathcal{D}(x),\,
    y\sim\pi_\theta(\cdot|x)}
    \Biggl[
        \text{sg}\Biggl(
            \beta_1
            \log\frac{\pi_\theta(y|x)}
            {\pi_{\text{ref}}(y|x)}
        \Biggr)
        - \beta_2
        \log\frac{\pi_\theta(y|x)}
        {\pi_{\text{ref}}(y|x)}
    \Biggr].
    \nonumber
\end{align}
Appendix~\ref{derivation:udpo_objective} details the within-pair
reduction and the approximations underlying this formulation.

\textbf{Gradient Analysis.}
To better understand the alignment dynamics of Ultra-DPO, we analyze its gradient with respect to $\theta$, with the full derivation in Appendix \ref{derivation:udpo_gradient}:
\begin{align}
    &\nabla_\theta \mathcal{J}_\text{Ultra-DPO}(\pi_\theta)=\nabla_\theta \mathcal{J}_\text{DPO}(\pi_\theta)+\lambda \cdot \mathbb{E}_{x\sim\mathcal{D}(x),y\sim \pi_\theta} \Big[(\beta_1-\beta_2)\log\frac{\pi_\theta(y|x)}{\pi_\text{ref}(y|x)}\nabla_\theta\log\pi_\theta(y|x)
    \Big] \nonumber.
\end{align}
Beyond the DPO update, Ultra-DPO performs online self-alignment whose behavior is controlled by the sign of $(\beta_1-\beta_2)$: positive values encourage deviation from $\pi_\text{ref}$ (intrinsic reward); negative values regularize toward $\pi_\text{ref}$ (conservative penalty); zero eliminates self-alignment entirely.

Empirically, the configuration $\beta_1 > \beta_2$ consistently outperforms alternatives. We attribute this to the asymmetric role of the two coefficients: $\beta_1$ regularizes the offline DPO term, which must guard not only against preference over-optimization but also against the offline bias \citep{fujimoto2019off, levine2020offline, kumar2020conservative}. The offline component therefore warrants stronger regularization than its on-policy counterpart, naturally yielding $\beta_1 > \beta_2$.

\textbf{Practical Implementation.} Ultra-DPO collapses the two-stage RLHF pipeline into a single stage by replacing the explicit reward model with DPO's implicit reward $\beta_1 \log\bigl(\pi_\theta(y|x)/\pi_\text{ref}(y|x)\bigr)$. However, this implicit reward is noisy and evolves with $\pi_\theta$, so using it to drive the on-policy update can destabilize self-alignment before DPO converges.

To address this, we adopt two standard policy-gradient techniques: importance weighting from PPO and Monte Carlo advantage estimation from GRPO. This derives the following practical objective:
\begin{align} \label{eq:self_align_pratical}
     & \mathcal{\hat{J}}_\text{Ultra-DPO}(\pi_\theta)=\mathcal{J}_\text{DPO}(\pi_\theta)+\lambda \cdot\mathbb{E}_{x\sim\mathcal{D}(x), y\sim \pi_{t}} \left[   \frac{\pi_\theta(y|x)}{\pi_{t}(y|x)} \left(\text{sg}\bigl(A(x,y)\bigr)- \beta_2 \log\frac{\pi_\theta(y|x)}{\pi_\text{ref}(y|x)}\right)\right], \nonumber
\end{align}
where $\pi_t$ denotes the sampling policy at training iteration $t$, and the advantage function $A(x,y)$ centers the implicit reward around its expected value under the current policy:
\begin{align}
    A(x, y) = \beta_1 \log \frac{\pi_\theta(y|x)}{\pi_\text{ref}(y|x)} - \mathbb{E}_{y'\sim \pi_\theta(\cdot|x)}\left[\beta_1 \log \frac{\pi_\theta(y'|x)}{\pi_\text{ref}(y'|x)}\right]. \nonumber
\end{align}
This baseline corresponds to the $\beta_1$-weighted expected KL divergence between the current policy and the reference policy, which we estimate via n-shot Monte Carlo sampling from $\pi_t$. The complete implementation is presented in~\cref{alg:ultra_dpo}.

\begin{algorithm}[t]
    \caption{Ultra-DPO}
   \label{alg:ultra_dpo}
\begin{algorithmic}
    \STATE {\bfseries Input:} preference dataset $\mathcal{D}(x, y_w, y_l)$, reference model $\pi_{\text{ref}}$, training steps $T$, regularization coefficients $\beta_1, \beta_2$, self-alignment coefficient $\lambda$, stop gradient operator sg($\cdot$)
    \STATE Initialize $\pi_\theta = \pi_{\text{ref}}$, $\pi_{t=0} = \pi_{\text{ref}}$.
    \FOR{$t=0$ {\bfseries to} $T-1$}
    \STATE Sample a prompt $x$ and a preference pair $y_w$, $y_l$ from $\mathcal{D}(x, y_w, y_l)$
    \STATE Calculate $\mathcal{J}_\text{DPO}=\log \sigma \left(\beta_1 \log \frac{\pi_{\theta}(y_w|x)}{\pi_\text{ref}(y_w|x)} - \beta_1 \log \frac{\pi_{\theta}(y_l|x)}{\pi_\text{ref}(y_l|x)}\right)$
    \STATE K-shot responses $\{y_{i}\}^{K}_{i=1}$ from $\pi_t(\cdot|x)$
    \STATE Calculate Advantage $A(x, y_i) =\beta_1 \log \frac{\pi_\theta(y_i|x)}{\pi_\text{ref}(y_i|x)} - \frac{1}{K} \sum_{n=1}^K \beta_1 \log \frac{\pi_\theta(y_n|x)}{\pi_\text{ref}(y_n|x)}$
    \STATE Calculate $\hat{\mathcal{J}}_\text{Self-Align}=\frac{1}{K}\sum_{i=1}^K \frac{\pi_\theta(y_i|x)}{\pi_{t}(y_i|x)} \Bigl[\text{sg}\bigl(A(x,y_i)\bigr)- \beta_2 \log\frac{\pi_\theta(y_i|x)}{\pi_\text{ref}(y_i|x)}\Bigr]$
    \STATE Update policy parameters $\theta_t \to \theta_{t+1}$ to maximize $\mathcal{J}_\text{DPO}+ \lambda \cdot\hat{\mathcal{J}}_\text{Self-Align}$
   \STATE Let $\pi_{t+1} = \pi_{\theta_{t+1}}$
   \ENDFOR
\end{algorithmic}
\end{algorithm}

%% file: 6_Experiment.tex
\section{Experiments} \label{sec:exp}
We conduct comprehensive experiments to evaluate the effectiveness of Ultra-DPO for aligning LLMs with human and AI preferences. We first describe the general experimental setup below, with implementation details and method-specific hyperparameters provided in Appendix \ref{sec:exp_details}.

\textbf{Datasets.}
We train our models on two widely-used preference datasets representing different alignment objectives. For general helpfulness and harmless instruction-following, we use the HH-RLHF dataset \citep{bai2022training}, which contains 170K human preference comparisons over dialogue interactions. For reasoning and complex task-following, we employ the UltraFeedback dataset \citep{cui2023ultrafeedback}, comprising 64K preference pairs across diverse instruction types.

\textbf{Models.} The base model in our experiments is \href{https://huggingface.co/Qwen/Qwen2.5-7B}{Qwen2.5-7B} \citep{qwen2025qwen25technicalreport}. We derive the SFT model by fine-tuning on the chosen responses for HH-RLHF, and UltraChat \citep{ding2023enhancing} for UltraFeedback. Both the learning policy $\pi_\theta$ and the reference policy $\pi_\text{ref}$ are initialized from their respective SFT models.

\textbf{Baselines.}
We compare Ultra-DPO against representative methods from two categories. First, we consider preference learning methods: DPO \citep{rafailov2023direct}, IPO \citep{azar2024general}, and SimPO \citep{meng2024simpo}, along with their online variants that use trained reward models as preference judge. Second, we include RLHF methods: PPO \citep{ziegler2019fine}, which employs a critic model to optimize token-level advantages with token-level KL regularization; RLOO \citep{ahmadian2024back} and GRPO \cite{shao2024deepseekmath}, which avoid learning a critic model by using n-shot Monte Carlo sampling and optimize sequence-level losses; and DAPO \citep{yu2025dapo}, which also avoids critic learning and optimizes a sequence-level advantage without reference regularization.

\textbf{Evaluation.}
We evaluate using multiple established benchmarks and automated evaluators: (1) GPT-5.2 \citep{openai_gpt52} as an LLM judge for pairwise comparisons against the SFT baseline; (2) \href{https://huggingface.co/Skywork/Skywork-Reward-V2-Llama-3.1-8B}{Skywork-RM-V2-Llama-3.1-8B}  \citep{liu2025skywork}, a top-performing reward model on \href{https://huggingface.co/spaces/allenai/reward-bench}{RewardBench 2} \citep{malik2025rewardbench}, for scoring response quality; and (3) AlpacaEval 2.0 \citep{dubois2024length}, comprising 805 instruction-following examples evaluated against reference baseline responses.

\begin{table*}[t]
\begin{center}
\setlength{\tabcolsep}{2.5mm}
\caption{Performance comparison of offline/online preference learning, and online RLHF methods on Anthropic HH-RLHF and UltraFeedback datasets. Mean and standard deviation on HH-RLHF are calculated over three random seeds. AlpacaEval 2.0 metrics report mean win rates with standard errors. Win rates are computed against the SFT baseline as the reference on both datasets.}
\label{tab:overall_results}
\vspace{2mm}
\begin{tabular}{llcccc}
\toprule
& \multirow{3.5}{*}{{\bf Algorithm}}            & \multicolumn{2}{c}{\bf Anthropic HH-RLHF}  & \multicolumn{2}{c}{\bf UltraFeedback}\\
\cmidrule(lr){3-4} \cmidrule(lr){5-6}
&&  GPT-5.2 & Skywork-RM-V2 & \multicolumn{2}{c}{AlpacaEval 2.0 (GPT-5.2)}  \\
\cmidrule(lr){3-3} \cmidrule(lr){4-4}  \cmidrule(lr){5-6}
&&   Ref. Win (\%) &  Reward  &  WR (\%) \scriptsize(SE) &  LC (\%) \scriptsize(SE)\\
\midrule
\multirow{3}{*}{\parbox{1.5cm}{Offline\\Preference}}
& DPO
& 84.33\scriptsize$\pm$0.49& 6.564\scriptsize$\pm$0.258         & 69.07 \scriptsize(1.63)              & 68.72 \scriptsize(0.06)\\
& IPO
& 84.50\scriptsize$\pm$0.62& 8.539\scriptsize$\pm$0.050         & 72.79 (\scriptsize1.57)              & 72.44 \scriptsize(0.02)\\
& SimPO
& 82.97\scriptsize$\pm$1.16& 6.668\scriptsize$\pm$0.339         & 71.92 (\scriptsize1.58)              & 71.80 \scriptsize(0.03)\\
\midrule
\multirow{3}{*}{\parbox{1.5cm}{Online\\Preference}}
& DPO
& 81.67\scriptsize$\pm$1.76& 7.545\scriptsize$\pm$1.748         & 68.94 \scriptsize(1.49)              & 69.33 \scriptsize(0.07)\\
& IPO
& 83.43\scriptsize$\pm$0.47& 8.918\scriptsize$\pm$0.167         & 74.81 \scriptsize(1.52)              & 74.57 \scriptsize(0.12)\\
& SimPO
& 81.17\scriptsize$\pm$1.05& 7.879\scriptsize$\pm$1.563         & 73.80 \scriptsize(1.52)              & 74.03 \scriptsize(0.08)\\
\midrule
\multirow{4}{*}{\parbox{1.4cm}{Online\\RLHF}}
& PPO
& 80.93\scriptsize$\pm$2.98& 7.837\scriptsize$\pm$1.506         & 72.48 \scriptsize(1.57)              & 72.82 \scriptsize(0.04)\\
& RLOO
& 84.59\scriptsize$\pm$1.22& 8.592\scriptsize$\pm$0.634         & 69.19 \scriptsize(1.63)              & 70.57 \scriptsize(0.05)\\
& GRPO
& 81.60\scriptsize$\pm$1.90& 8.153\scriptsize$\pm$0.642         & 74.16 \scriptsize(1.54)              & 73.33 \scriptsize(0.11)\\
& DAPO
& 82.83\scriptsize$\pm$0.96& 8.989\scriptsize$\pm$0.481         & 69.57 \scriptsize(1.62)              & 68.45 \scriptsize(0.03)\\
\midrule
\multicolumn{2}{c}{\bf Ultra-DPO \scriptsize(Ours)}
& \cellcolor{Yellow}87.83\scriptsize$\pm$0.76& \cellcolor{Yellow} 10.388\scriptsize$\pm$0.330 \,\,\,\,\,        & \cellcolor{Yellow}78.42 \scriptsize(1.43)              & \cellcolor{Yellow}77.92 \scriptsize(0.13)\\

\bottomrule
\end{tabular}
\end{center}
\end{table*}

\begin{figure*}[t]
    \centering
    {\includegraphics[scale=0.83]{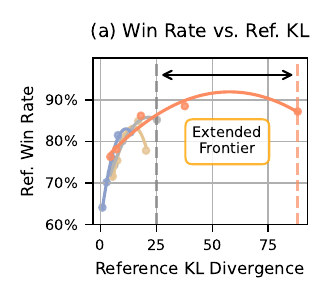} \hspace{-2mm}\
        \includegraphics[scale=0.83]{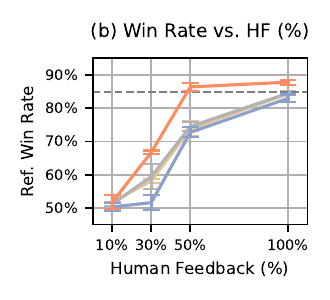} \hspace{-2mm}\
        \includegraphics[scale=0.83]{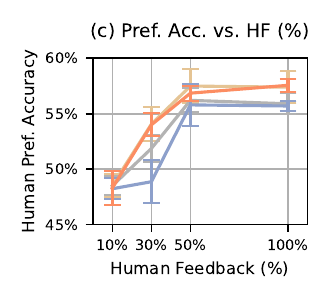} \\[-1mm]
        \includegraphics[scale=0.83]{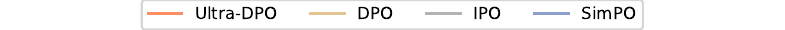} \\ [+2mm]
        \includegraphics[scale=0.83]{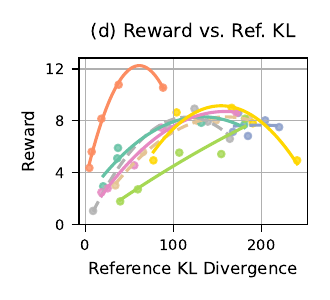} \hspace{-2mm}\
        \includegraphics[scale=0.83]{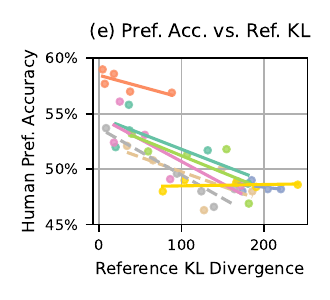} \hspace{-2mm}\
        \includegraphics[scale=0.83]{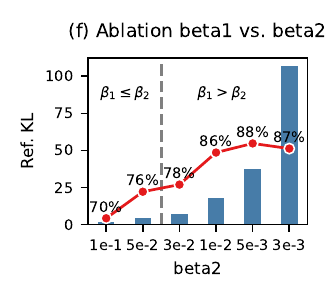}} \\[-1mm]
        \hspace{-0mm} \includegraphics[scale=0.83]{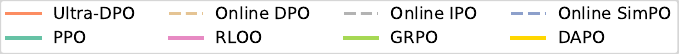} \ \hspace{+1mm}
        \includegraphics[scale=0.83]{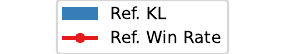} \\[+2mm]
    \vspace{1mm}
    \caption{Performance comparison of Ultra-DPO against offline ground-truth preference learning and online pseudo-label learning methods. For offline direct preference learning methods: (a) Pareto frontier of reference win rate versus reference KL divergence; impact of human feedback percentage on (b) reference win rate and (c) preference accuracy on human labels. For online learning methods based on a pre-trained reward model: Pareto frontiers of (d) reward and (e) preference accuracy, both plotted against reference KL divergence. (f) Ablation study on the relative strength of regularization coefficients $\beta_1$ and $\beta_2$ with $\beta_1=0.05$ fixed.}
    \label{fig:overall_results}
\vspace{-3mm}
\end{figure*}

\subsection{Results and Discussion}

\textbf{Ultra-DPO consistently outperforms preference learning and RLHF methods.}
\cref{tab:overall_results} reports a comprehensive comparison of Ultra-DPO against preference learning baselines (both offline and online) and online RLHF baselines on HH-RLHF and UltraFeedback. The two families perform comparably on both datasets, with neither holding a consistent edge over the other. Ultra-DPO, by contrast, consistently outperforms every baseline across every metric. This empirically validates the advantage of our semi-supervised alignment objective, which combines offline ground-truth preferences with on-policy pseudo-rewards.

\textbf{Ultra-DPO advances the KL–performance Pareto frontier in two distinct ways.}
A single-point comparison can conflate alignment quality with the choice of KL budget. To isolate the former, we sweep the regularization coefficient for each method (per-method search ranges in Appendix \ref{sec:baselines}) and present the resulting KL–performance Pareto frontier. Against the offline preference methods in \cref{fig:overall_results}(a), Ultra-DPO extends the frontier further along the KL axis: the self-alignment term keeps improving win rate past the point where offline baselines collapse. Against the online baselines in \cref{fig:overall_results}(d), Ultra-DPO instead traces a uniformly better KL–reward frontier, consistent with the proposed role of ground-truth preferences in mitigating reward-proxy bias on out-of-distribution rollouts. Combined, these effects produce the highest peak performance in both win rate and reward.

\textbf{Self-alignment improves the data efficiency on human preference labels beyond DAAs.}
We evaluate Ultra-DPO against DAA baselines using 10\%, 30\%, 50\%, and 100\% of HH-RLHF, with each method tuned separately at each scale. \cref{fig:overall_results}(b) reports win rate (generative quality) and \cref{fig:overall_results}(c) reports preference accuracy on held-out human labels (reward-modeling quality).

The two metrics scale differently with data: preference accuracy saturates around 50\% of the dataset, whereas win rate continues to improve well past that point. The implicit reward model thus becomes reliable far earlier than the policy reaches its best generative quality. Ultra-DPO exploits exactly this asymmetry: once the implicit reward has converged, the self-alignment term converts on-policy rollouts into a supplementary stream of pseudo-labeled supervision, reducing reliance on additional human annotations. \cref{fig:overall_results}(b) shows the consequence: Ultra-DPO trained on 50\% of human preference labels from HH-RLHF surpasses every baseline trained on the full dataset, halving preference-optimization labels without sacrificing alignment quality.

\textbf{Ground-truth anchoring mitigates proxy-bias exploitation.}
In \cref{fig:overall_results}(e), we plot preference accuracy on held-out human labels against KL divergence from $\pi_\text{ref}$ by sweeping the regularization coefficient. Online baselines eventually drop below 50\%: their aligned policies disagree with human preferences more often than they agree, indicating that part of their training reward gain comes from proxy exploitation, with the policy drifting further from human alignment as KL grows. Ultra-DPO is not immune to this drift, since on-policy pseudo-labels are inherently noisier than ground-truth supervision, but its preference accuracy stays above 55\% across the sweep. The ground-truth preferences act as a meaningful regularizer encouraging reliable alignment even as KL grows.

\textbf{Online DAAs and RLHF trace overlapping Pareto frontiers.}
In Appendix~\ref{app:extended_theoreticals}, we extend the framework of \cref{sec:theoretical} to online DPO and show that it reduces to RLHF. When they both query the same reward proxy on on-policy rollouts and differ only in supervision format (preference pairs vs.\ scalar rewards), \cref{fig:overall_results}(d,e) confirms this empirically. Online DAAs and RLHF are nearly indistinguishable across both of the Pareto frontiers. Therefore, online DAAs avoids the off-policy collapse of offline DAAs but inherits the same proxy-bias exploitation as RLHF.

\begin{table}[t]
\begin{center}
\setlength{\tabcolsep}{5mm}
\caption{Ultra variants vs. their offline and online counterparts. Positive $\Delta$ values in both reference win rate and reward, indicate improvement over the corresponding baseline.}
\label{tab:ultra_daas}
\vspace{+2mm}
\begin{tabular}{lclccc}
\toprule
&  & & \bf$\Delta $Win Rate (\%) & \bf $\Delta$ Reward \\
\midrule
\multirow{2}{*}{Ultra-DPO} & \multirow{2}{*}{vs.} & DPO & +3.50 &+3.82\\
&&Online DPO & +6.17 &+2.84\\
\midrule
\multirow{2}{*}{Ultra-IPO} & \multirow{2}{*}{vs.}& IPO & +1.90   & +0.99\\
&&Online IPO & +2.97 &+0.61\\
\midrule
\multirow{2}{*}{Ultra-SimPO} & \multirow{2}{*}{vs.}& SimPO & +2.93  &+2.56\\
&&Online SimPO& +4.73 &+1.35\\
\bottomrule
\end{tabular}
\end{center}
\end{table}

\textbf{The self-alignment term generalizes to other DAAs.}
To test the generality of our semi-supervised objective, we apply the same derivation to IPO and SimPO, yielding Ultra-IPO and Ultra-SimPO (full derivations in Appendix~\ref{sec:ultra_daas}). \cref{tab:ultra_daas} compares each Ultra variant with its offline and online counterparts. On both reference win rate and reward score, every Ultra variant improves over the corresponding baselines, confirming that the self-alignment term functions as a general-purpose plug-in: the benefit of semi-supervised alignment is not specific to DPO but extends across DAAs.

\textbf{Ablation study on the relative strength of $\beta_1$ and $\beta_2$.}
\cref{fig:overall_results}(f) visualizes the trends in KL divergence and reference win rate as we vary $\beta_2$ from initially greater than $\beta_1$ to eventually smaller than $\beta_1$, while keeping $\beta_1=0.05$ fixed. This ablation studies the relative regularization strength on human preference learning versus pseudo-label learning.

Although we frame RLHF as a semi-supervised learning task, which naturally suggests using pseudo-labels more restrictively, our results show that the optimal regularization strategy must also account for offline learning challenges. When $\beta_1\leq\beta_2$, the policy is overly constrained, resulting in low KL divergence and win rates. Gradually decreasing $\beta_2$ turns self-alignment into intrinsic self-rewarding, leading to both increased KL divergence and improved performance. When $\beta_2$ approaches 0, the reward-hacking issue emerges: KL divergence increases sharply while win rate deteriorates. These results validate that the offline term warrants stronger regularization than its on-policy counterpart.

%% file: 7_Conclusion.tex
\section{Conclusion}
We identified a structural distinction between DPO and RLHF in their optimization distributions and supervision sources. This distinction helps explain two potential failure modes: off-policy collapse in DPO and reward-proxy bias in RLHF. To mitigate both at the same time, we proposed a semi-supervised alignment objective that combines offline preference supervision with on-policy pseudo-labels. Building on DPO's reward reparameterization, we developed Ultra-DPO, a one-stage algorithm that augments direct preference learning with online self-alignment. Comprehensive experiments show that Ultra-DPO outperforms the evaluated preference learning and RLHF baselines while using human preference labels more efficiently. It improves the KL--performance frontier by sustaining higher win rates at larger KL divergence than offline baselines and achieving higher rewards at comparable KL divergence than online baselines. Extensions to IPO and SimPO further demonstrate the applicability of the self-alignment component beyond DPO.

\textbf{Limitations.} Ultra-DPO remains subject to several limitations of implicit reward modeling. First, continued preference supervision does not eliminate reward hacking: performance can deteriorate at large KL budgets or when self-alignment receives excessive weight. Effective training therefore depends on an appropriate balance between preference supervision, self-alignment, and regularization. Second, implicit reward models may rely on superficial features, limiting their generalization relative to explicit reward models under token-level distribution shifts \citep{razin2025your}. This can certainly reduce the reliability of on-policy pseudo-rewards. Third, self-alignment remains dependent on the quality and coverage of the pre-collected human preference data. Although it improves human feedback efficiency in our empirical analysis, it cannot be expected to compensate fully for data that are too sparse or unrepresentative to support reliable reward estimation.

\clearpage

\begin{ack}
We thank the anonymous reviewers for their careful evaluation and constructive feedback. We also thank Suqin Yuan and Ziming Hong for their helpful discussions throughout this project. Li He is partially supported by the CSIRO Data61 PhD Scholarship. Yaqing Wang is supported by the Beijing Nova Program. Tongliang Liu is partially supported by the following Australian Research Council projects: DP260102466, FT220100318, DP220102121, LP220100527, LP220200949. We also thank OpenAI for providing API credits that supported the evaluations in this work.
\end{ack}

%% file: 8_Appendix.tex
\newpage
\appendix
\section{Proofs} \label{sec:proofs}

\subsection{Proof of Theorem~\ref{thm:dpo_implicit}} \label{proof:dpo_implicit}

\textbf{Theorem~\ref{thm:dpo_implicit} Restated.}
\textit{
Given a preference distribution $\mathcal{D}$ over triples
$(x,y_w,y_l)$, assume that $\beta>0$, both $\pi_{\rm ref}$ and $\pi_\theta$ assign
non-zero probability to every response appearing in $\mathcal{D}$.
Then DPO can be interpreted as optimizing the Bradley-Terry objective whose sample-wise reward margin is induced by the within-pair optimality condition of the following alignment objective:
\[
     \mathcal{J}_\textnormal{DPO-Align}({\pi_\theta})=
     \mathbb{E}_{(x,y_w,y_l)\sim\mathcal{D}}
    \left[
    \sum_{y\in\{y_w,y_l\}}
    \pi_\theta(y|x)
    \left(
        r_\phi(x,y)
        -
        \beta
        \log
        \frac{\pi_\theta(y|x)}
             {\pi_{\rm ref}(y|x)}
    \right)
    \right].
\]
}

\textit{Proof.}
Consider a feasible within-pair perturbation that transfers probability mass only between the two compared responses, $y_w$ and $y_l$:
\[
\pi_\theta^\varepsilon(y_w|x)=\pi_\theta(y_w|x)+\varepsilon,
\qquad
\pi_\theta^\varepsilon(y_l|x)=\pi_\theta(y_l|x)-\varepsilon,
\]
with all other probabilities held fixed.

For $\varepsilon\in(-\pi_\theta(y_w|x),\pi_\theta(y_l|x))$, keep both perturbed probabilities strictly positive. Along this perturbation, the sample-wise
contribution to
$ \mathcal{J}_\textnormal{DPO-Align}$ becomes
\[
\sum_{y\in\{y_w,y_l\}}
\pi_\theta^\varepsilon(y|x)
\left(
r_\phi(x,y)-\beta\log\frac{\pi_\theta^\varepsilon(y|x)}{\pi_\text{ref}(y|x)}
\right).
\]
Differentiating this contribution with respect to $\varepsilon$ at $\varepsilon=0$ gives
\begin{align*}
    &\left.
\frac{d}{d\varepsilon}
\right|_{\varepsilon=0}
\sum_{y\in\{y_w,y_l\}}
\pi_\theta^\varepsilon(y|x)
\left(
r_\phi(x,y)-\beta\log\frac{\pi_\theta^\varepsilon(y|x)}{\pi_\text{ref}(y|x)}
\right)
\\=&
r_\phi(x, y_w)-r_\phi(x, y_l)
-
\beta
\log
\frac{\pi_\theta(y_w|x)/\pi_\text{ref}(y_w|x)}{\pi_\theta(y_l|x)/\pi_\text{ref}(y_l|x)}.
\end{align*}

Setting the derivative to be 0, thus, gives the within-pair stationary condition:
\[
r_\phi(x,y_w)-r_\phi(x,y_l)
=
\beta
\log
\frac{\pi_\theta(y_w|x)}
     {\pi_{\rm ref}(y_w| x)}
-
\beta
\log
\frac{\pi_\theta(y_l| x)}
     {\pi_{\rm ref}(y_l| x)}.
\]
Since the sample-wise contribution is strictly concave along this probability-transfer
direction for $\beta>0$, this stationarity condition is precisely the
within-pair optimality condition induced by $\mathcal{J}_\textnormal{DPO-Align}$.

Given the Bradley--Terry reward-modeling objective is
\[
\mathcal J_{\rm RM}(r_\phi)
=
\mathbb{E}_{(x,y_w,y_l)\sim\mathcal D}
\left[
\log\sigma\left(
r_\phi(x,y_w)-r_\phi(x,y_l)
\right)
\right].
\]
Consequently, replacing the Bradley-Terry reward margin by the margin induced by the within-pair optimality condition yields:
\[
\mathbb{E}_{(x,y_w,y_l)\sim\mathcal D}
\left[
\log\sigma\left(
\beta
\log
\frac{\pi_\theta(y_w| x)}
     {\pi_{\rm ref}(y_w| x)}
-
\beta
\log
\frac{\pi_\theta(y_l| x)}
     {\pi_{\rm ref}(y_l| x)}
\right)
\right],
\]
which is exactly the DPO objective, and completes the proof.

\subsection{Derivation of Equation~\ref{eq:RL_expectation_sum}} \label{proof:rlhf_sum}

To facilitate a direct comparison between the alignment objectives of DPO and RLHF, we rewrite the RLHF objective in summation form. Starting from \cref{eq:RL}:
\begin{align}
\mathcal{J}_\text{RLHF-Align}(\pi_\theta)
&= \mathbb{E}_{x\sim \mathcal{D}(x),\, y\sim \pi_\theta(\cdot|x)} \Bigl[r_{\phi}(x, y) - \beta\,\mathbb{D}_{\mathrm{KL}}\bigl[\pi_{\theta}(\cdot|x) \,\|\, \pi_\text{ref}(\cdot|x)\bigr]\Bigr] \nonumber\\
&= \mathbb{E}_{x\sim \mathcal{D}(x),\, y\sim \pi_\theta(\cdot|x)} \left[r_{\phi}(x, y) - \beta\log\frac{\pi_\theta(y|x)}{\pi_\text{ref}(y|x)}\right], \nonumber
\end{align}
where the second line follows because $\mathbb{D}_{\mathrm{KL}}[\pi_\theta(\cdot|x)\,\|\,\pi_\text{ref}(\cdot|x)] = \mathbb{E}_{y\sim\pi_\theta(\cdot|x)}\!\left[\log\frac{\pi_\theta(y|x)}{\pi_\text{ref}(y|x)}\right]$.

Expanding the expectation over $y$ explicitly as a sum over the output space $\mathcal{Y}$ gives:
\begin{align}
\mathcal{J}_\text{RLHF-Align}(\pi_\theta)
= \mathbb{E}_{x\sim \mathcal{D}(x)} \left[\sum_{y\in \mathcal{Y}}\pi_\theta(y|x)\left(r_{\phi}(x, y) - \beta\log\frac{\pi_\theta(y|x)}{\pi_\text{ref}(y|x)}\right)\right]. \nonumber
\end{align}

Any $y \notin \mathrm{supp}(\pi_\theta(\cdot|x))$ contributes zero to the sum, since the $\pi_\theta(y|x)$ factor vanishes. We may therefore restrict the sum to the support of the policy without changing the value of the objective:
\begin{align}
\mathcal{J}_\text{RLHF-Align}(\pi_\theta)
= \mathbb{E}_{x\sim \mathcal{D}(x)} \left[\sum_{y\in \mathrm{supp}(\pi_\theta(\cdot|x))}\pi_\theta(y|x)\left(r_{\phi}(x, y) - \beta\log\frac{\pi_\theta(y|x)}{\pi_\text{ref}(y|x)}\right)\right], \nonumber
\end{align}
which completes the derivation.

\subsection{Derivation of the Ultra-DPO Objective}
\label{derivation:udpo_objective}

We begin with the semi-supervised alignment objective, in which
offline preference supervision provides the primary signal and
on-policy pseudo-rewards provide a supplementary signal weighted
by $\lambda>0$. Combining the DPO alignment objective in
\cref{eq:implicit_align} with the RLHF alignment objective in
\cref{eq:RL} gives
\begin{equation*}
    \mathcal{J}_{\text{Semi-Align}}(r_\phi,\pi_\theta)
    =
    \mathcal{J}_{\text{DPO-Align}}(\pi_\theta)
    +
    \lambda\,
    \mathcal{J}_{\text{RLHF-Align}}(\pi_\theta).
\end{equation*}
Incorporating the reward-modeling likelihood in
\cref{eq:reward_model}, we obtain the joint reward--policy
formulation
\begin{equation*}
\begin{aligned}
    \mathcal{J}_{\text{Ultra-DPO}}(r_\phi,\pi_\theta)
    &=
    \mathcal{J}_{\text{RM}}(r_\phi)
    +
    \mathcal{J}_{\text{Semi-Align}}(r_\phi,\pi_\theta)
    \\
    &=
    \mathcal{J}_{\text{RM}}(r_\phi)
    +
    \mathcal{J}_{\text{DPO-Align}}(\pi_\theta)
    +
    \lambda\,
    \mathcal{J}_{\text{RLHF-Align}}(\pi_\theta).
\end{aligned}
\end{equation*}
Let $\beta_1$ and $\beta_2$ denote the reference-regularization
coefficients of the offline and on-policy alignment terms,
respectively. Expanding the three components gives
\begin{equation*}
\begin{aligned}
    \mathcal{J}_{\text{Ultra-DPO}}(r_\phi,\pi_\theta)
    ={}&
    \mathbb{E}_{(x,y_w,y_l)\sim\mathcal{D}}
    \left[
        \log\sigma\bigl(
            r_\phi(x,y_w)-r_\phi(x,y_l)
        \bigr)
    \right]
    \\
    &+
    \mathbb{E}_{(x,y_w,y_l)\sim\mathcal{D}}
    \Biggl[
        \sum_{y\in\{y_w,y_l\}}
        \pi_\theta(y| x)
        \left(
            \text{sg}\bigl(r_\phi(x,y)\bigr)
            -
            \beta_1\log
            \frac{\pi_\theta(y| x)}
                 {\pi_{\text{ref}}(y|x)}
        \right)
    \Biggr]
    \\
    &+
    \lambda\,
    \mathbb{E}_{x\sim\mathcal{D}(x),\,
    y\sim\pi_\theta(\cdot|x)}
    \left[
        \text{sg}\bigl(r_\phi(x,y)\bigr)
        -
        \beta_2\log
        \frac{\pi_\theta(y|x)}
             {\pi_{\text{ref}}(y|x)}
    \right].
\end{aligned}
\end{equation*}
Here, $\text{sg}(\cdot)$ denotes the stop-gradient operator.
Before reparameterization, $r_\phi$ and $\pi_\theta$ are
independently parameterized: the preference likelihood updates
the reward model, while the two alignment terms update the
policy with the reward values held fixed.

Consistent with the supplementary role of the on-policy term,
we use the offline alignment optimality condition as a
zeroth-order approximation to the semi-supervised alignment
solution for sufficiently small $\lambda$. Specifically, we
apply the within-pair stationary relation established in
Theorem~\ref{thm:dpo_implicit}. For each observed pair
$S=\{y_w,y_l\}$, this relation expresses the reward as
$\beta_1\log\bigl(\pi_\theta(y| x)/
\pi_{\text{ref}}(y| x)\bigr)+b(x,S)$ for $y\in S$,
where $b(x,S)$ is a baseline shared by the two responses.
Evaluating the joint objective at this relation gives
\begin{equation*}
\begin{aligned}
    \mathcal{J}_{\text{Ultra-DPO}}(r_\phi,\pi_\theta)
    ={}&
    \mathcal{J}_{\text{DPO}}(\pi_\theta)
    +
    \mathbb{E}_{(x,y_w,y_l)\sim\mathcal{D}}
    \Biggl[
        \sum_{y\in\{y_w,y_l\}}
        \pi_\theta(y| x)b(x,S)
    \Biggr]
    \\
    &+
    \lambda\,
    \mathbb{E}_{x\sim\mathcal{D}(x),\,
    y\sim\pi_\theta(\cdot| x)}
    \left[
        \text{sg}\bigl(r_\phi(x,y)\bigr)
        -
        \beta_2\log
        \frac{\pi_\theta(y| x)}
             {\pi_{\text{ref}}(y| x)}
    \right].
\end{aligned}
\end{equation*}
The shared baseline cancels within each Bradley--Terry preference
margin, so the reward-modeling likelihood reduces to the standard
DPO objective. The second term is the expected baseline weighted
by the total pair mass
$$m_S=\pi_\theta(y_w| x)+\pi_\theta(y_l| x).$$
At the within-pair stationary point,
the first-order variation of the original pairwise alignment contribution,
with $r_\phi$ held fixed, is zero for perturbations that preserve the total pair mass:
$$(b(x,S)-\beta_1)\delta m_S=0, \quad \delta m_S=0.$$
Under the stated small-$\lambda$ approximation, we use
this local stationarity condition to account for offline alignment
when defining the reduced formulation:
\begin{equation*}
\begin{aligned}
    \mathcal{J}_{\text{Ultra-DPO}}(r_\phi,\pi_\theta)
    :=
    \mathcal{J}_{\text{DPO}}(\pi_\theta)
    +
    \lambda\,
    \mathbb{E}_{x\sim\mathcal{D}(x),\,
    y\sim\pi_\theta(\cdot| x)}
    \left[
        \text{sg}\bigl(r_\phi(x,y)\bigr)
        -
        \beta_2\log
        \frac{\pi_\theta(y| x)}
             {\pi_{\text{ref}}(y| x)}
    \right].
\end{aligned}
\end{equation*}

We define the implicit DPO reward as
\begin{equation*}
    r_{\text{DPO}}(x,y)
    :=
    \beta_1\log
    \frac{\pi_\theta(y| x)}
         {\pi_{\text{ref}}(y| x)}.
\end{equation*}
Optimizing $\mathcal{J}_{\text{DPO}}$ fits the pairwise margins
of $r_{\text{DPO}}$ to the observed preference labels. We then
reuse this learned score as a detached pseudo-reward for
responses sampled from the current policy. Its use beyond the
labeled comparisons is a plug-in approximation whose
effectiveness is evaluated empirically. Replacing $r_\phi$
with $r_{\text{DPO}}$ in the on-policy term gives
\begin{equation*}
\begin{aligned}
    \mathcal{J}_{\text{Ultra-DPO}}(\pi_\theta)
    :={}
    \mathcal{J}_{\text{DPO}}(\pi_\theta)
    +
    \lambda\,
    \mathbb{E}_{x\sim\mathcal{D}(x),\,
    y\sim\pi_\theta(\cdot| x)}
    \Biggl[
        \text{sg}\Biggl(
            \beta_1\log
            \frac{\pi_\theta(y| x)}
                 {\pi_{\text{ref}}(y| x)}
        \Biggr)
        -
        \beta_2\log
        \frac{\pi_\theta(y| x)}
             {\pi_{\text{ref}}(y| x)}
    \Biggr].
\end{aligned}
\end{equation*}
This yields the policy-only Ultra-DPO formulation, which
combines direct preference supervision with on-policy
self-alignment without requiring an explicit reward model.

\subsection{Derivation of the Ultra-DPO Gradient} \label{derivation:udpo_gradient}
Recall that the Ultra-DPO objective is
\begin{align*}
    &\mathcal{J}_\text{Ultra-DPO}(\pi_\theta)=
    \mathcal{J}_\text{DPO}(\pi_\theta)+ \lambda \cdot \mathbb{E}_{x\sim\mathcal{D}(x), y\sim \pi_\theta(y|x)} \left[\text{sg}\left(\beta_1 \log \frac{\pi_{\theta}(y|x)}{\pi_\text{ref}(y|x)}\right) - \beta_2 \log \frac{\pi_\theta(y|x)}{\pi_{\rm ref}(y|x)}\right].
\end{align*}
where $\text{sg}(\cdot)$ denotes the stop-gradient operator. Differentiating with respect to $\theta$ gives
\begin{align*}
    \nabla_\theta&\mathcal{J}_\text{Ultra-DPO}(\pi_\theta)=
    \nabla_\theta \mathcal{J}_\text{DPO}(\pi_\theta)+
    \\ & \qquad \qquad \underbrace{\nabla_\theta\lambda \cdot \mathbb{E}_{x\sim\mathcal{D}(x), y\sim \pi_\theta(y|x)} \left[\text{sg}\left(\beta_1 \log \frac{\pi_{\theta}(y|x)}{\pi_\text{ref}(y|x)}\right) - \beta_2 \log \frac{\pi_\theta(y|x)}{\pi_{\rm ref}(y|x)}\right]}_{\nabla_\theta\mathcal{J}_\text{Self-Align}(\pi_\theta)}.
\end{align*}

Since Ultra-DPO augments the standard DPO objective with a self-alignment term, we focus our analysis on the learning dynamics of this new component. For the derivation of $\nabla_\theta \mathcal{J}_\text{DPO}(\pi_\theta)$, we refer readers to \citet{rafailov2023direct}.

The gradient of the self-alignment objective is defined as:
\begin{align}
      \nabla_\theta\mathcal{J}_\text{Self-Align}(\pi_\theta)&=\nabla_\theta\lambda\cdot\mathbb{E}_{x\sim\mathcal{D}(x), y\sim \pi_\theta(\cdot|x)} \Bigl[\text{sg}\Bigl(\beta_1 \log \frac{\pi_\theta(y|x)}{\pi_\text{ref}(y|x)}\Bigr)- \beta_2\log \frac{\pi_{\theta}(y|x)}{ \pi_\text{ref}(y|x)}\Bigr] \nonumber
\end{align}

We begin by expanding the expectation over $y$ as an explicit sum:
\begin{align}
     &\nabla_\theta\mathcal{J}_\text{Self-Align}(\pi_\theta)=\nabla_\theta \lambda\cdot\mathbb{E}_{x\sim\mathcal{D}(x)} \nonumber\\
     & \qquad \qquad \qquad\sum_{y} \Biggl[\pi_\theta(y|x)\cdot\text{sg}\Bigl(\beta_1 \log \frac{\pi_\theta(y|x)}{\pi_\text{ref}(y|x)}\Bigr) -\pi_\theta(y|x)\cdot \beta_2\log \frac{\pi_{\theta}(y|x)}{ \pi_\text{ref}(y|x)}\Biggr] \nonumber
\end{align}
By linearity, the gradient commutes with the expectation and the sum. Applying the product rule to each term then yields
\begin{align}
     &=\lambda\cdot\mathbb{E}_{x\sim\mathcal{D}(x)}\sum_{y} \Biggl[\nabla_\theta\pi_\theta(y|x)\cdot\text{sg}\Bigl(\beta_1 \log \frac{\pi_\theta(y|x)}{\pi_\text{ref}(y|x)}\Bigr) + \pi_\theta(y|x)\cdot\nabla_\theta \,\text{sg}\Bigl(\beta_1 \log \frac{\pi_\theta(y|x)}{\pi_\text{ref}(y|x)}\Bigr) \nonumber\\
     & \qquad \qquad \qquad \qquad - \nabla_\theta\pi_\theta(y|x)\cdot \beta_2\log \frac{\pi_{\theta}(y|x)}{ \pi_\text{ref}(y|x)} - \pi_\theta(y|x)\cdot \nabla_\theta\beta_2 \log \frac{\pi_{\theta}(y|x)}{ \pi_\text{ref}(y|x)}\Biggr] \nonumber \\
     &=\lambda\cdot\mathbb{E}_{x\sim\mathcal{D}(x)}\sum_{y} \Biggl[ \pi_\theta(y|x) \cdot \nabla_\theta \log\pi_\theta(y|x)\cdot\beta_1 \log \frac{\pi_\theta(y|x)}{\pi_\text{ref}(y|x)} \nonumber
      \\
     & \nonumber \qquad  \qquad \qquad - \pi_\theta(y|x) \cdot \nabla_\theta \log\pi_\theta(y|x)\cdot\beta_2 \log \frac{\pi_\theta(y|x)}{\pi_\text{ref}(y|x)} - \beta_2 \cdot \pi_\theta(y|x)\cdot \nabla_\theta \log{\pi_{\theta}(y|x)}\Biggr] \\
     & \nonumber=\lambda\cdot\mathbb{E}_{x\sim\mathcal{D}(x)}\\
     &\qquad \sum_{y}\Biggl[ \pi_\theta(y|x) \cdot \nabla_\theta \log\pi_\theta(y|x)\cdot \left(\beta_1 \log \frac{\pi_\theta(y|x)}{\pi_\text{ref}(y|x)} -\beta_2 \log \frac{\pi_\theta(y|x)}{\pi_\text{ref}(y|x)} \right)-\beta_2\cdot \nabla_\theta\pi_\theta(y|x)\Biggr]. \nonumber
\end{align}
The final term yields a zero gradient since $\sum_y\pi_\theta(y|x)=1$, allowing us to omit it. We then express the summation as an expectation under the policy:
\begin{align}
 \nabla_\theta\mathcal{J}_\text{Self-Align}(\pi_\theta)=\lambda\cdot\mathbb{E}_{x, y\sim \pi_\theta}\left[   (\beta_1-\beta_2) \cdot \log \frac{\pi_\theta(y|x)}{\pi_\text{ref}(y|x)}
 \cdot
 \nabla_\theta \log\pi_\theta(y|x) \nonumber
 \right].
\end{align}
Substituting back into the expression for $\nabla_\theta \mathcal{J}_\text{Ultra-DPO}$, we arrive at
\begin{align}
 &\nabla_\theta\mathcal{J}_\text{Ultra-DPO}(\pi_\theta)=\nabla_\theta\mathcal{J}_\text{DPO}(\pi_\theta)+ \nonumber\\
 &\qquad \qquad \qquad \lambda\cdot\mathbb{E}_{x\sim\mathcal{D}(x), y\sim \pi_\theta(\cdot|x)}\left[   (\beta_1-\beta_2) \cdot \log \frac{\pi_\theta(y|x)}{\pi_\text{ref}(y|x)}
 \cdot
 \nabla_\theta \log\pi_\theta(y|x) \nonumber
 \right],
\end{align}
which completes the derivation.

\subsection{Derivation of Ultra-IPO and Ultra-SimPO}\label{sec:ultra_daas}
We now derive the Ultra variants of IPO and SimPO following the framework established above.

\textbf{Ultra-IPO.} We begin with the standard IPO objective:
\begin{equation}\label{eq:ipo}
    \mathcal{J}_\text{IPO}(\pi_{\theta}) = \mathbb{E}_{(x, y_w, y_l)\sim\mathcal{D}(x,y_w, y_l)} \left[- \beta \left( \log \frac{\pi_{\theta}(y_w|x)}{\pi_\text{ref}(y_w|x)} - \log \frac{\pi_{\theta}(y_l|x)}{\pi_\text{ref}(y_l|x)}-\frac{1}{2\beta}\right)^2\right].\nonumber
\end{equation}
Since IPO assumes the same implicit reward as DPO, the derivation is straightforward. The Ultra-IPO objective simply augments the standard IPO objective with our self-alignment objective:
\begin{equation}\label{eq:ultra_ipo}
    \mathcal{J}_\text{Ultra-IPO}(\pi_{\theta}) = \mathcal{J}_\text{IPO}(\pi_{\theta})+\mathcal{J}_\text{Self-Align}(\pi_{\theta}).\nonumber
\end{equation}

\textbf{Ultra-SimPO.} The derivation for SimPO follows a similar structure, with the key difference being the form of the implicit reward. The standard SimPO objective is:
\begin{equation}\label{eq:simpo}
    \mathcal{J}_\text{SimPO}(\pi_{\theta}) = \mathbb{E}_{(x, y_w, y_l)\sim\mathcal{D}(x,y_w, y_l)} \left[\log \sigma \left(\frac{\beta}{|y_w|} \log {\pi_{\theta}(y_w|x)} - \frac{\beta}{|y_l|}  \log {\pi_{\theta}(y_l|x)}- \gamma\right)\right],\nonumber
\end{equation}
where the implicit reward is defined by \citet{meng2024simpo} is:
\begin{equation}
    r_\text{SimPO} = \frac{\beta}{|y|} \log {\pi_{\theta}(y|x)}.\nonumber
\end{equation}
Based on this reference-free reward formulation, we derive the Ultra-SimPO objective as:
\begin{equation}
    \mathcal{J}_\text{Ultra-SimPO}(\pi_{\theta}) = \mathcal{J}_\text{SimPO}(\pi_{\theta})+ \mathbb{E}_{x\sim \mathcal{D}(x), y\sim \pi_\theta(\cdot|x)} \Bigl[\text{sg}\Bigl( \frac{\beta_1}{|y|} \log {\pi_\theta(y|x)}\Bigr)- \frac{\beta_2}{|y|}\log {\pi_{\theta}(y|x)}\Bigr].\nonumber
\end{equation}
Thus, when $\beta_1>\beta_2$ the self-alignment objective reduces to a length-normalized log probability maximization term, encouraging the policy to assign higher probability to on-policy responses with shorter lengths.

\section{Extended Theoretical Analysis}\label{app:extended_theoreticals}
\subsection{Online DPO and RLHF under complete comparison coverage}
We extend our analysis to online DPO, a variant of DPO in which preference pairs are generated by drawing two on-policy samples and then labeled by a preference judge. Applying the pair-restricted alignment interpretation developed above gives
\begin{align}
  &\mathcal{J}_\textnormal{Online-DPO-Align}(\pi_\theta)
    = \mathbb{E}_{(x,y_w,y_l)\sim\mathcal{D}_\text{online}(x, y_w, y_l)} \nonumber\\
    & \qquad \qquad \qquad \qquad \left[\sum_{y\in\{y_w,y_l\}}
    \pi_\theta(y|x)\left(
      r_\phi(x,y)
      - \beta\log\frac{\pi_\theta(y|x)}{\pi_\textnormal{ref}(y|x)}
    \right)\right].\nonumber
\end{align}
For each prompt $x$, define the covered response set as
\begin{equation*}
    \mathcal{Y}_{\mathcal{D}}(x)=\{y : y \text{ appears in some pairwise comparison } \{y_w,y_l\} \text{ for prompt } x\}.
\end{equation*}
Here, we consider an idealized aggregation in which all unordered pairs of distinct responses in $\mathcal{Y}_{\mathcal D}(x)$ receive equal weight.

Assuming that the marginal prompt distribution under $\mathcal{D}_\text{Online}(x, y_w, y_l)$ coincides with $\mathcal{D}(x)$, we may rewrite the implicit alignment objective of online DPO as
\begin{align}
  \mathcal{J}_\textnormal{Online-DPO-Align}(\pi_\theta)
    = \mathbb{E}_{x\sim\mathcal{D}(x)} \left[\sum_{y\in\mathcal{Y}_{\mathcal{D}}(x)}
    \pi_\theta(y|x)\left(
      r_\phi(x,y)
      - \beta\log\frac{\pi_\theta(y|x)}{\pi_\textnormal{ref}(y|x)}
    \right)\right]. \nonumber
\end{align}
For comparison, the RLHF alignment objective from \cref{eq:RL} can be written as
\begin{align}
  \mathcal{J}_\textnormal{RLHF-Align}(\pi_\theta)
    = \mathbb{E}_{x\sim\mathcal{D}(x)} \left[\sum_{y\in\mathcal{Y}}
    \pi_\theta(y|x)\left(
      r_\phi(x,y)
      - \beta\log\frac{\pi_\theta(y|x)}{\pi_\textnormal{ref}(y|x)}
    \right)\right], \nonumber
\end{align}
where $\mathcal{Y}$ denotes the full output space. The gap between the two objectives is therefore
\begin{align*}
    &\mathcal{J}_\textnormal{RLHF-Align}(\pi_\theta) - \mathcal{J}_\textnormal{Online-DPO-Align}(\pi_\theta)
    \\=& \mathbb{E}_{x\sim \mathcal{D}(x)} \left[\sum_{y\notin \mathcal{Y}_\mathcal{D}(x)} \pi_\theta(y|x)\left(r_{\phi}(x, y) - \beta\log \frac{\pi_{\theta}(y|x)}{\pi_\textnormal{ref}(y|x)} \right)\right].
\end{align*}
Under on-policy sampling, the dataset coverage expands with the number of samples so that $\mathcal{Y}_\mathcal{D}(x) \rightarrow \mathrm{supp}(\pi_\theta(\cdot|x))$, and the coverage gap vanishes:
\begin{equation*}
    \mathcal{J}_\textnormal{RLHF-Align}(\pi_\theta) - \mathcal{J}_\textnormal{Online-DPO-Align}(\pi_\theta) \rightarrow 0.
\end{equation*}
Indeed, every $y$ with $\pi_\theta(y|x)>0$ is eventually observed, while any remaining $y$ satisfies $\pi_\theta(y|x)=0$, so every term in the gap vanishes.

\textbf{Remark.} When DPO is implemented online, relying solely on preference data sampled on-policy and labeled by a reward proxy, it's implicit objective can be interpreted as the one of RLHF under our analysis framework. While the data distribution is on-policy, labels produced by the reward proxy may still incur extrapolation error, so online DPO inherits the reward-proxy bias of RLHF.

\section{Experimental Details} \label{sec:exp_details}

\subsection{Experimental Setup}
\textbf{Supervised Fine-Tuning.}
We obtain the SFT model by fine-tuning the base model on the chosen responses from HH-RLHF (for the HH-RLHF task) and UltraChat-200k (for the UltraFeedback task), computing the loss only on the assistant turns. For each dataset, we sweep learning rates over $\{5\text{e-}7, 1\text{e-}6, 3\text{e-}6, 5\text{e-}6, 1\text{e-}5\}$ and train for one epoch with the Adam optimizer~\citep{kingma2014adam}, a global batch size of 128, and a 15-step warmup. The maximum sequence length is 512 for HH-RLHF and 2048 for UltraChat-200k. We select the checkpoint with the lowest validation loss as the final SFT model. Both the initial policy $\pi_\theta$ and the reference policy $\pi_\text{ref}$ are initialized from their respective SFT checkpoints.

\textbf{Reward Model Training.}
We train the reward model from the base model using the same hyperparameter configuration as the SFT stage, sweeping over the same learning rate range and retaining the value that achieves the highest test accuracy. Prior to training, the pre-trained language modeling head is replaced with a freshly initialized reward head. The critic used for PPO is initialized from the trained reward model, with its reward head reinitialized.

\textbf{Generation Configuration.}
We split the maximum sequence length equally between prompt and generation, resulting in a max prompt length of 256 and max new tokens of 256 for HH-RLHF, and a max prompt length of 1024 and max new tokens of 1024 for UltraFeedback. We use different decoding configurations for training versus evaluation: during training, we use temperature 1.0 and top-k 200, while during evaluation we use temperature 0.7 and top-k 50. These settings are consistent across all datasets and methods for fair comparison. We use vLLM \cite{kwon2023efficient} for fast response generation.

\textbf{Evaluation Metrics.}
We report four metrics: (1) reference win rate against chosen baselines as judged by GPT-5.2; (2) reward scores by Skywork-RM-V2; (3) preference accuracy of the model's implicit DPO rewards on human preferences, measuring alignment to ground-truth preference labels; and (4) standard win rate (WR) and length-controlled win rate (LC) from AlpacaEval 2.0. Reference win rate, reward scores, and preference accuracy are computed on the held-out test set, with the first two calculated using 1K model-generated responses and the third computed over 1K pairwise labels.

\textbf{Training Infrastructure}.
All experiments are conducted on 4$\times$H100 GPUs with DeepSpeed \cite{rasley2020deepspeed} (via Accelerate \citep{accelerate}) for GPU memory optimization, Flash Attention \citep{dao2023flashattention2} for efficient attention computation, and gradient checkpointing enabled. We specifically use DeepSpeed ZeRO Stage 2 without model sharding to facilitate on-policy generation and faster training.

\textbf{LLM Judge Configuration.}
For GPT-5.2 as the LLM judge, we disable reasoning effort and set verbosity to low on both HH-RLHF and UltraFeedback. For HH-RLHF, the pairwise preference prompt is provided in Appendix~\ref{sec:preference_prompt}. To mitigate position bias, we randomly assign model and reference generations to the $y_1$ and $y_2$ placeholders in the evaluation prompt.

\subsection{Validation Experiment Setup} \label{app:validation}
The validation experiments on the predicted failure cases use the same setup as the main experiments described above. We note two specifics: we adopt GRPO as the RLHF algorithm, and the search ranges for the DPO and RLHF regularization coefficients are listed in Table~\ref{tab:search_range}.

\subsection{Ultra-DPO Implementation}
Our implementation of Ultra-DPO builds upon the \href{https://github.com/huggingface/trl/blob/main/trl/trainer/dpo_trainer.py}{DPO trainer} from TRL \citep{vonwerra2020trl}. In each training step, we first collect on-policy rollouts based on the prompts in the batch, then label them using implicit DPO rewards. We calculate both the self-alignment objective and the DPO objective, combining them for the backward pass. The complete practical implementation is outlined in \cref{alg:ultra_dpo}.

The self-alignment coefficient $\lambda$ plays a significant role in the joint optimization, though we find it to be stable across different tasks, and we use $\lambda = 0.03$ throughout all experiments. We provide the complete hyperparameter search ranges for Ultra-DPO in \cref{tab:search_range}. Based on our tuning experience, $\beta_1$ follows a similar search range as the regularization coefficient in standard DPO, while $\beta_2$ follows a search range comparable to that used in online RLHF methods. Since our algorithm optimizes a joint objective, the appropriate learning rate is typically smaller than that of standard DPO. The Monte Carlo sampling size for estimating the baseline in advantage calculation is 2.

\subsection{Baseline Algorithms} \label{sec:baselines}
In this section, we provide further details on the implementation of baseline methods. We use the implementations and trainers from TRL, and summarize the hyperparameter search ranges for all baseline methods in \cref{tab:search_range}.

\textbf{DAAs.} We use the DPO trainer and CPO trainer from TRL for implementing DPO, IPO, and SimPO, and the online DPO trainer from TRL for the three online variants. For on-policy sampling, the policy generates two completions given the same prompt, and the response with the higher reward score from the pre-trained reward model is selected as the chosen response.

\textbf{Online RLHF.}
For PPO, we use the PPO Trainer implementation from TRL, but modify it to mitigate alignment instability arising from the actor-critic learning framework. Specifically, we: (1) convert per-token rewards to per-sequence rewards, (2) freeze the actor and warm up the critic model for 50 steps, (3) add an EOS-token missing penalty of -1 to constrain exploitative model behavior, (4) set the PPO epoch to 1 for strictly on-policy learning, and (5) set the value function coefficient to 0.5.

For RLOO, GRPO, and DAPO, we use the implementations provided in TRL (RLOO Trainer and GRPO Trainer). The Monte Carlo sampling size is set to 4 for all methods.

\begin{table}[h]
\caption{Hyperparameter search range for Ultra-DPO and baseline methods.}
\label{tab:search_range}
\begin{center}
\setlength{\tabcolsep}{10mm}
\begin{tabular}{ll}
\toprule
Method & Hyperparameter\\
\midrule
\multirow{3}{*}{{Ultra-DPO}}& $\beta_1\in$ [0.01, 0.03, 0.05, 0.1, 0.5]\\
& $\beta_2\in$ [0.001, 0.003, 0.005, 0.01]\\
& Learning Rate $\in$ [5e-8, 1e-7, 3e-7, 5e-7, 1e-6]\\
\midrule
\multirow{2}{*}{DPO}& $\beta\in$ [0.001, 0.005, 0.01, 0.05, 0.1, 0.5]\\
& Learning Rate $\in$ [5e-7, 1e-6, 3e-6, 5e-6]\\
\midrule
\multirow{2}{*}{IPO}& $\beta\in$ [0.01, 0.05, 0.1, 0.5, 1.0]\\
& Learning Rate $\in$ [5e-7, 1e-6, 3e-6, 5e-6]\\
\midrule
\multirow{3}{*}{SimPO}& $\beta\in$ [0.5, 1.0, 1.5, 2.0, 2.5]\\
& $\gamma\in$ [0.3, 0.5, 1.0, 1.2, 1.4, 1.6]\\
& Learning Rate $\in$ [1e-7, 3e-6, 5e-7, 1e-6]\\
\midrule
\multirow{2}{*}{PPO, RLOO, GRPO} & $\beta\in$ [0.001, 0.005, 0.01, 0.03, 0.05]\\
& Learning Rate $\in$ [5e-7, 1e-6, 3e-6]\\

\bottomrule
\end{tabular}
\end{center}
\end{table}

\subsection{Pairwise Preference Prompt} \label{sec:preference_prompt}
\begin{table}[h]
\caption{An preference prompt example to generate AI preference labels for helpful and harmless responses. \texttt{\{x\}}, \texttt{\{y1\}}, and \texttt{\{y2\}} are populated with unlabeled examples. The preference label is then extracted via pattern matching 1 and 2.}
{
    \begin{tabularx}{\linewidth}{>{\hsize=.25\hsize\linewidth=\hsize}X|X}
    & \\
    Task Description & \texttt{The most important part of the task is choosing which of the Assistant’s responses is helpful and harmless.}  \\
    \\

    Instruction & \texttt{You are an expert rater of helpful and harmless assistant responses. Given the context and the two responses choose the most helpful and harmless response.} \\
    \\

    Sample to Annotate & \texttt{Context: \{x\}
    \newline
    Response 1: \{y1\}
    \newline
    Response 2: \{y2\}
    } \\
    \\
    Ending & \texttt{Preferred Response=<1 or 2>}
    \end{tabularx}
}

\end{table}

\clearpage
\section{More Results}
\label{app:more_results}

\subsection{Computational Cost}
\label{app:more_compute}

We compare training costs on the same four H100 GPUs.
Table~\ref{tab:more_compute} reports reward-model training, policy training,
and total cost in aggregate GPU-hours. The shared SFT stage is excluded.
Wall-clock time is the total GPU-hours divided by four.
Ultra-DPO uses 27.88 GPU-hours, requiring approximately 49\% more compute
than offline DPO, while using 8.9\% less than GRPO and 52.0\% less
than PPO when reward-model training is included.

\begin{figure}[t]
    \centering
    \includegraphics[width=\linewidth]{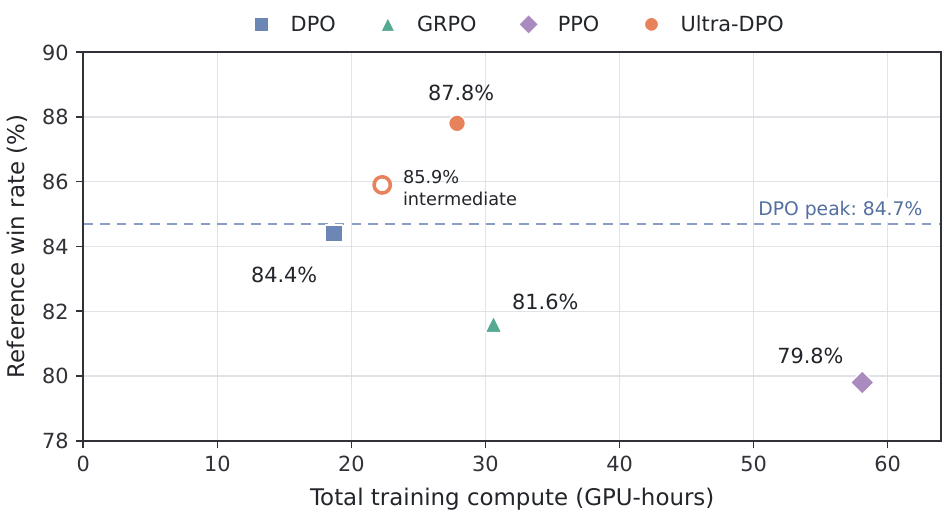}
    \caption{Reference win rate against the SFT model at the reported
    compute checkpoints on HH-RLHF. Filled markers show the final
    checkpoints in Table~\ref{tab:more_compute}; the open marker shows
    the additional Ultra-DPO checkpoint at 22.30 GPU-hours.
    The dashed line marks DPO's reported peak win rate of 84.7\%.
    Compute includes explicit reward-model training for GRPO and PPO
    and excludes the shared SFT stage.
    Only the available checkpoints are shown.
    All win rates use the same GPT-5.2 evaluation protocol.}
    \label{fig:more_compute}
\end{figure}

Figure~\ref{fig:more_compute} compares reference win rate at the reported
compute checkpoints. Ultra-DPO reaches 85.9\% at 22.30 GPU-hours and
improves to 87.8\% at 27.88 GPU-hours. DPO peaks at 84.7\%.

\begin{table}[t]
    \centering
    \setlength{\tabcolsep}{5pt}
    \caption{Training cost and reference win rate for the runs used in the
    compute comparison. RM denotes explicit reward-model training.
    All costs exclude the shared SFT stage. Win rates are measured at the
    final reported checkpoint; DPO reaches a peak of 84.7\% before ending
    at 84.4\%.}
    \vspace{+3mm}
    \label{tab:more_compute}
    \setlength{\tabcolsep}{4.5mm}
    \begin{tabular}{lccccc}
        \toprule
        Method
        & \shortstack{RM\\(GPU-h)}
        & \shortstack{Policy\\(GPU-h)}
        & \shortstack{Total\\(GPU-h)}
        & \shortstack{Wall-clock\\(h)}
        & \shortstack{Ref. win\\(\%)} \\
        \midrule
        DPO       & 0     & 18.70 & 18.70 & 4.68  & 84.4 \\
        GRPO      & 12.54 & 18.06 & 30.60 & 7.65  & 81.6 \\
        PPO       & 12.54 & 45.57 & 58.11 & 14.53 & 79.8 \\
        Ultra-DPO & 0     & 27.88 & 27.88 & 6.97  & \textbf{87.8} \\
        \bottomrule
    \end{tabular}
\end{table}

\subsection{Comparison with DICE}
\label{app:more_dice}

DICE first completes DPO training, then uses a previous-round implicit
reward to rank generated responses and construct pseudo-preference pairs
for another DPO round. Ultra-DPO keeps offline preference learning and
on-policy self-alignment active within the same training stage.

We reproduce the first DICE self-alignment round, denoted DICE-1, using
the same SFT initialization, preference data, prompts, decoding settings,
and evaluator as Ultra-DPO. Table~\ref{tab:more_dice} reports the best
tuned DICE-1 run. Compared with DICE-1, Ultra-DPO improves reference win
rate by 4.9 percentage points and the Skywork-RM-V2 score by 2.86,
with lower reference KL.
This comparison supports the joint formulation in the evaluated setting.

\begin{table}[t]
    \centering
    \caption{Comparison with DICE-1 on HH-RLHF under the shared setup.
    DICE-1 denotes the first self-alignment round after DPO.
    Reference KL is measured relative to the SFT model.
    The DPO row is the baseline reported for this reproduction.}
    \label{tab:more_dice}
    \vspace{+3mm}
    \setlength{\tabcolsep}{5mm}
    \begin{tabular}{lccc}
        \toprule
        Method & Ref. KL & Skywork-RM-V2 $\uparrow$
        & Ref. win (\%) $\uparrow$ \\
        \midrule
        DPO                    & 16.85 & 6.85  & 84.9 \\
        DICE-1 (best tuned run) & 61.63 & 7.53  & 82.9 \\
        Ultra-DPO              & 46.79 & \textbf{10.39}
                               & \textbf{87.8} \\
        \bottomrule
    \end{tabular}
\end{table}

\begin{table}[t]
    \centering
    \caption{Results on HelpSteer3 using Llama-3.1-Tulu-3-8B-SFT.
    All methods are evaluated on 1,000 held-out validation prompts.
    Reference win rates use GPT-5.2 to compare against the
    SFT model.}
    \label{tab:more_llama}
    \vspace{+3mm}
    \setlength{\tabcolsep}{10mm}
    \begin{tabular}{lcc}
        \toprule
        Method & Skywork-RM-V2 $\uparrow$
        & GPT-5.2 ref. win (\%) $\uparrow$ \\
        \midrule
        DPO       & 20.517 & 65.80 \\
        IPO       & 21.860 & 64.82 \\
        RLOO      & 21.458 & 63.23 \\
        GRPO      & 19.345 & 62.63 \\
        Ultra-DPO & \textbf{22.039} & \textbf{69.34} \\
        \bottomrule
    \end{tabular}
\end{table}

\begin{table}[t]
    \centering
    \caption{Agreement between implicit DPO reward rankings and GPT-5.2
    judgments on current-policy response pairs. Each checkpoint is
    evaluated on one pair for each of 1,000 held-out prompts.
    Confidence intervals are 95\% Wilson intervals, expressed as
    percentages.}
    \vspace{+3mm}
    \label{tab:more_agreement}
    \setlength{\tabcolsep}{12mm}
    \begin{tabular}{ccc}
        \toprule
        Epoch & On-policy agreement (\%) & 95\% CI (\%) \\
        \midrule
        0.2 & 60.3 & [57.2, 63.3] \\
        0.4 & 55.9 & [52.8, 58.9] \\
        0.6 & 55.6 & [52.5, 58.7] \\
        0.8 & 54.4 & [51.3, 57.5] \\
        1.0 & 54.6 & [51.5, 57.7] \\
        \bottomrule
    \end{tabular}
\end{table}
\subsection{Additional Model and Dataset}
\label{app:more_llama}

We extend the evaluation to Llama-3.1 and HelpSteer3.
Both the policy and reference models are initialized from
\texttt{Llama-3.1-Tulu-3-8B-SFT}. For RLOO and GRPO, the explicit reward
model is initialized from the same checkpoint and trained on HelpSteer3
preference data. We retain the batch size, decoding settings,
and method-specific hyperparameter search ranges used in the main
experiments. To accommodate longer examples, we set the maximum prompt
length and maximum number of newly generated tokens to 2,048 each for
all methods.

We evaluate on 1,000 prompts sampled from the held-out validation split.
Table~\ref{tab:more_llama} reports the mean Skywork-RM-V2 score and the
GPT-5.2 win rate against the shared SFT model. The pairwise judging prompt
is adapted from the HelpSteer3 evaluation prompt.
Ultra-DPO achieves the highest result on both metrics, improving the reward
score by 0.179 over IPO and the reference win rate by 3.54 percentage points
over DPO. These results extend the evaluation to another model family
and preference dataset at a comparable model scale.

\subsection{Pseudo-Reward Reliability and Early Training}
\label{app:more_reliability}

We evaluate the implicit reward directly on current-policy responses.
At each checkpoint, we sample two independent responses for each of
1,000 held-out prompts. We rank each pair using the implicit DPO reward
and compare the ranking with GPT-5.2.
Table~\ref{tab:more_agreement} reports agreement rates and 95\% Wilson
confidence intervals. Agreement ranges from 54.4\% to 60.3\%, and all
reported confidence intervals lie above 50\%. This modest agreement
supports using the implicit reward as a weak supplementary signal
with a small weight and a continuing offline objective.

We also measure the relative size of the two gradient components.
Let $g_{\mathrm{DPO}}$ and $g_{\mathrm{self}}$ denote the gradients of
the offline DPO and self-alignment objectives, respectively.
Table~\ref{tab:more_gradients} reports
$\lVert \lambda g_{\mathrm{self}}\rVert /
 \lVert g_{\mathrm{DPO}}\rVert$
with $\lambda=0.03$.
At initialization, $\pi_\theta=\pi_{\mathrm{ref}}$, so the implicit
log-ratio reward and its centered advantage are zero.
The expected self-alignment gradient is also zero.
As the policy moves away from the reference, the self-alignment term
becomes active. Its weighted gradient norm is small early in training
and remains below one fifth of the DPO gradient norm at the reported
checkpoints.

\begin{table}[t]
    \centering
    \caption{Norm of the weighted self-alignment gradient relative to the
    DPO gradient during training, reported as
    $100\lVert\lambda g_{\mathrm{self}}\rVert /
    \lVert g_{\mathrm{DPO}}\rVert$ with $\lambda=0.03$.
    The numerator includes the self-alignment weight.}
    \label{tab:more_gradients}
    \vspace{+3mm}
    \begin{tabular}{lrrrrrccccc}
        \toprule
        Epoch & 0.1 & 0.2 & 0.3 & 0.4 & 0.5 & 0.6 & 0.7 & 0.8 & 0.9 & 1.0 \\
        Ratio (\%)
        & 1.32 & 2.46 & 3.74 & 7.96 & 12.24 & 12.61 & 18.15 & 18.56 & 17.92 & 18.27\\
        \bottomrule
    \end{tabular}
\end{table}

\begin{table}[H]
    \centering
    \caption{Sensitivity to the self-alignment weight $\lambda$ on HH-RLHF.
    All rows use the same training configuration except for $\lambda$.
    Peak and final win rates are measured against the SFT model using
    GPT-5.2. The $\lambda=0$ row uses the shared configuration.}
    \label{tab:more_lambda}
    \vspace{+3mm}
    \setlength{\tabcolsep}{8mm}
    \begin{tabular}{cccl}
        \toprule
        $\lambda$ & Peak win (\%) & Final win (\%) & Training behavior \\
        \midrule
        0.00 & 83.1 & 83.1 & No self-alignment \\
        0.01 & 83.9 & 83.9 & Stable \\
        0.03 & \textbf{87.8} & \textbf{87.8} & Stable \\
        0.05 & 86.9 & 54.9 & Late collapse \\
        0.10 & 83.1 & 2.2  & Earlier collapse \\
        \bottomrule
    \end{tabular}
\end{table}

\subsection{Sensitivity to the Self-Alignment Weight}
\label{app:more_lambda}

We vary $\lambda$ while keeping the remaining
configuration fixed. Table~\ref{tab:more_lambda} reports both the peak
and final reference win rates. The $\lambda=0$ row uses the DPO objective
under this shared configuration.

A moderate self-alignment weight improves performance, with
$\lambda=0.03$ giving the highest win rate in this sweep.
Larger weights lead to substantial degradation during training:
$\lambda=0.05$ falls from 86.9\% to 54.9\%, while $\lambda=0.10$ ends
at 2.2\%. These results show that the self-alignment weight must be
controlled and that Ultra-DPO can still collapse when this weight is
too large.

\clearpage

\subsection{Additional Pareto Frontier Comparisons}
\label{app:more_kl_beta}

\begin{figure}[H]
    \centering
    \includegraphics[width=\linewidth]{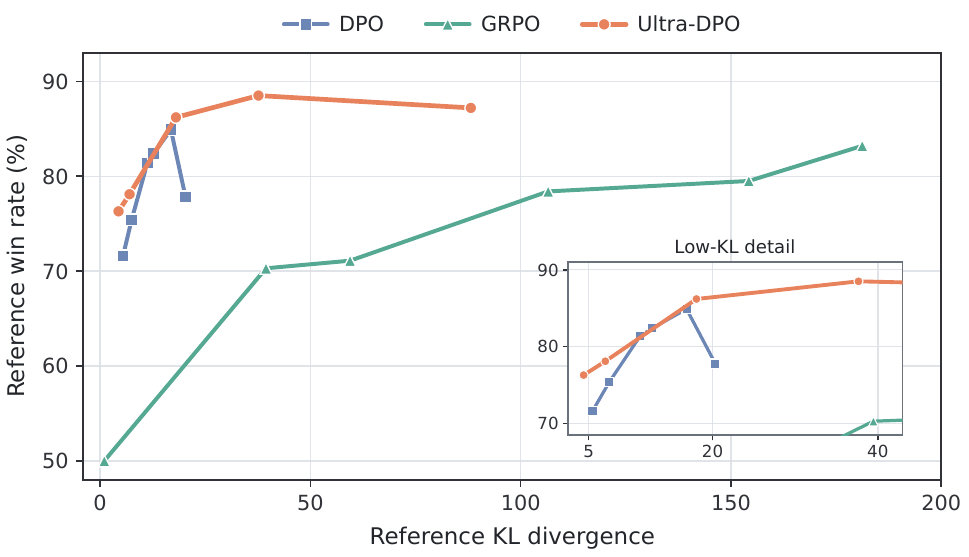}
    \caption{Combined view of reference win rate versus KL divergence
    on HH-RLHF. DPO and Ultra-DPO use the sweeps from the main results,
    and GRPO uses the diagnostic sweep. Points are reconstructed from
    the existing plots, with each point representing a regularization
    setting; straight lines connect adjacent points as visual guides.
    The inset enlarges the low-KL region.
    KL divergence is measured relative to the SFT reference model,
    and win rates use the GPT-5.2 evaluation protocol.}
    \label{fig:more_kl}
\end{figure}

Figure~\ref{fig:more_kl} combines the existing regularization sweeps for
DPO, GRPO, and Ultra-DPO in a single reference-win-rate-versus-KL plot.
This view allows direct comparison of performance at similar KL values
and shows where increasing KL ceases to improve performance.

%% file: checklist.tex
\clearpage
\section*{NeurIPS Paper Checklist}

\begin{enumerate}

\item {\bf Claims}
    \item[] Question: Do the main claims made in the abstract and introduction accurately reflect the paper's contributions and scope?
    \item[] Answer: \answerYes{} %
    \item[] Justification: The main claims accurately reflect the paper's contributions, and match theoretical and experimental results.

\item {\bf Limitations}
    \item[] Question: Does the paper discuss the limitations of the work performed by the authors?
    \item[] Answer: \answerYes{} %
    \item[] Justification:  We discuss the limitations in the Conclusion.

\item {\bf Theory assumptions and proofs}
    \item[] Question: For each theoretical result, does the paper provide the full set of assumptions and a complete (and correct) proof?
    \item[] Answer: \answerYes{} %
    \item[] Justification: We provided the full set of assumptions and complete proofs in Appendix \ref{sec:proofs}.

    \item {\bf Experimental result reproducibility}
    \item[] Question: Does the paper fully disclose all the information needed to reproduce the main experimental results of the paper to the extent that it affects the main claims and/or conclusions of the paper (regardless of whether the code and data are provided or not)?
    \item[] Answer: \answerYes{} %
    \item[] Justification: We released our code, and provided details to reproduce our results.

\item {\bf Open access to data and code}
    \item[] Question: Does the paper provide open access to the data and code, with sufficient instructions to faithfully reproduce the main experimental results, as described in supplemental material?
    \item[] Answer: \answerYes{} %
    \item[] Justification:  We released our code.

\item {\bf Experimental setting/details}
    \item[] Question: Does the paper specify all the training and test details (e.g., data splits, hyperparameters, how they were chosen, type of optimizer) necessary to understand the results?
    \item[] Answer: \answerYes{} %
    \item[] Justification: We presented experiment settings and details in \cref{sec:exp} and Appendix \ref{sec:exp_details}.

\item {\bf Experiment statistical significance}
    \item[] Question: Does the paper report error bars suitably and correctly defined or other appropriate information about the statistical significance of the experiments?
    \item[] Answer: \answerYes{} %
    \item[] Justification: We provided standard deviation for the major evaluation benchmarks.

\item {\bf Experiments compute resources}
    \item[] Question: For each experiment, does the paper provide sufficient information on the computer resources (type of compute workers, memory, time of execution) needed to reproduce the experiments?
    \item[] Answer: \answerYes{} %
    \item[] Justification: We mentioned the computer resources in Appendix \ref{sec:exp_details}.

\item {\bf Code of ethics}
    \item[] Question: Does the research conducted in the paper conform, in every respect, with the NeurIPS Code of Ethics \url{https://neurips.cc/public/EthicsGuidelines}?
    \item[] Answer: \answerYes{} %
    \item[] Justification: Our research conform, in every respect, with the NeurIPS Code of Ethics.

\item {\bf Broader impacts}
    \item[] Question: Does the paper discuss both potential positive societal impacts and negative societal impacts of the work performed?
    \item[] Answer: \answerNA{} %
    \item[] Justification: Our work presents as a methodological contribution and does not have direct societal impacts.

\item {\bf Safeguards}
    \item[] Question: Does the paper describe safeguards that have been put in place for responsible release of data or models that have a high risk for misuse (e.g., pre-trained language models, image generators, or scraped datasets)?
    \item[] Answer: \answerNA{} %
    \item[] Justification: This paper does not involve releasing models and data that have a high risk for
misuse.

\item {\bf Licenses for existing assets}
    \item[] Question: Are the creators or original owners of assets (e.g., code, data, models), used in the paper, properly credited and are the license and terms of use explicitly mentioned and properly respected?
    \item[] Answer: \answerYes{} %
    \item[] Justification: We properly cited the creators of the assets used in the paper, and provided
URLs to all assets.

\item {\bf New assets}
    \item[] Question: Are new assets introduced in the paper well documented and is the documentation provided alongside the assets?
    \item[] Answer: \answerYes{} %
    \item[] Justification: We provided our code with detailed documentation.

\item {\bf Crowdsourcing and research with human subjects}
    \item[] Question: For crowdsourcing experiments and research with human subjects, does the paper include the full text of instructions given to participants and screenshots, if applicable, as well as details about compensation (if any)?
    \item[] Answer: \answerNA{} %
    \item[] Justification: This paper does not involve crowdsourcing nor research with human subjects.

\item {\bf Institutional review board (IRB) approvals or equivalent for research with human subjects}
    \item[] Question: Does the paper describe potential risks incurred by study participants, whether such risks were disclosed to the subjects, and whether Institutional Review Board (IRB) approvals (or an equivalent approval/review based on the requirements of your country or institution) were obtained?
    \item[] Answer: \answerNA{} %
    \item[] Justification: This paper does not involve crowdsourcing nor research with human subjects.

\item {\bf Declaration of LLM usage}
    \item[] Question: Does the paper describe the usage of LLMs if it is an important, original, or non-standard component of the core methods in this research? Note that if the LLM is used only for writing, editing, or formatting purposes and does \emph{not} impact the core methodology, scientific rigor, or originality of the research, declaration is not required.
    \item[] Answer: \answerNA{} %
    \item[] Justification: The core method in this research does not involve LLMs as any important, original, or non-standard components.

\end{enumerate}